\documentclass[11pt]{article}

\usepackage{acl}

\usepackage{times}
\usepackage{latexsym}
\usepackage{amssymb}
\usepackage[T1]{fontenc}

\usepackage[utf8]{inputenc}

\usepackage{microtype}

\usepackage{inconsolata}

\usepackage{graphicx}
\usepackage{multirow}
\usepackage{amsmath}
\usepackage{tikz}
\usetikzlibrary{calc,decorations.text}

\usepackage{svg}
\usepackage{xcolor}
\usetikzlibrary{calc}

\usepackage{booktabs}
\usepackage{multirow}

\usepackage{xcolor}
\usepackage[most]{tcolorbox}
\usepackage[table]{xcolor}
\usepackage{tabularx}
\usepackage{subcaption}
\usepackage{booktabs}
\usepackage{array}
\usepackage{arydshln}
\usepackage{makecell}
\usepackage{placeins}
\usepackage{pgfplots}
\usetikzlibrary{matrix}
\pgfplotsset{compat=1.18}

\usepgfplotslibrary{fillbetween}
\usetikzlibrary{intersections}

\definecolor{highcol}{RGB}{184,220,174}
\definecolor{midcol}{RGB}{173,199,233}
\definecolor{lowcol}{RGB}{236,158,158}
\definecolor{memoryblue}{RGB}{70,110,230}

\definecolor{memoryblue}{RGB}{242,248,255} 
\definecolor{promptblue}{RGB}{217,225,242}   
\definecolor{promptline}{RGB}{90,90,90}      

\definecolor{highcol}{RGB}{191,217,198}   
\definecolor{midcol}{RGB}{172,198,228}    
\definecolor{lowcol}{RGB}{234,182,176}    
\definecolor{accentcol}{RGB}{88,126,184}  
\definecolor{linecol}{RGB}{85,85,85}      

\definecolor{qblue}{RGB}{55,126,184}
\definecolor{qorange}{RGB}{230,159,0}
\definecolor{qgreen}{RGB}{0,158,115}
\definecolor{qgray}{RGB}{120,120,120}

\definecolor{midgray}{gray}{0.35}
\definecolor{lightgray}{gray}{0.62}

\newcommand{\MCFrames}{%
4+
\textcolor{midgray}{8}+
\textcolor{lightgray}{32}%
}
\newcommand{\mcblue}[1]{\cellcolor{memoryblue}#1}
\def\method{{\textsc{MemoryCard}}}
\title{\method{}: Topic-Aware Multi-Modal Clue Compression for Long-Video Question Answering}

\author{Qing Yang$^{1}$, Pengcheng Huang$^{1}$, Xinze Li$^{1}$, Zhenghao Liu$^{1}$\thanks{ \ \ indicates corresponding author.}, \\ \textbf{Yukun Yan$^{2}$, Yu Gu$^{1}$, Ge Yu$^{1}$, Gang Li$^{3}$, Maosong Sun$^{2,4}$} \\ 
$^1$School of Computer Science and Engineering, Northeastern University, Shenyang, China \\
$^2$Department of Computer Science and Technology, Institute for AI, Tsinghua University, China\\
$^3$Digital China Group, Beijing, China \\
$^4$Jiangsu Collaborative Innovation Center for Language Ability, China\\
}

\begin{document}
\maketitle

\begin{abstract}
Long-video question answering remains challenging for Vision-Language Models (VLMs), as answer-relevant evidence is often sparse, transient, and temporally dispersed across lengthy video contexts. Existing frame-centric approaches improve efficiency through uniform sampling, query-aware frame selection, visual-token compression, and adaptive resolution strategies. However, they still rely on isolated and fragmented frames as the fundamental evidence units, limiting VLMs' ability to effectively capture coherent event-level semantics.
To address this limitation, we propose \method{}, a video-memory-based augmentation framework that organizes long videos into self-contained \textit{Memory Cards}. Specifically, \method{} first performs a self-reading process over videos and aligned utterances to segment the video into semantically coherent units, each corresponding to a distinct topic or event. For each unit, it generates an event-level video gist and selects representative visual moments, which are then rendered into unified Memory Cards for retrieval and question answering. Experimental results demonstrate that \method{} consistently improves long-video QA performance under comparable visual-token budgets, achieving up to a 21.8\% relative improvement in accuracy. All code is available at https://github.com/NEUIR/MemoryCard.

\end{abstract}


\section{Introduction}

Vision-Language Models (VLMs) have achieved remarkable progress in video understanding, enabling models to perceive dynamic visual content and answer questions about videos~\cite{DBLP:conf/emnlp/ZhangLB23,DBLP:conf/icml/ShenX0W0ZLXVBL025}. Recent advances have further improved the long-video understanding capabilities of VLMs through video instruction tuning and temporal-aware modeling~\cite{DBLP:conf/acl/0001RKK24,DBLP:conf/cvpr/RenYL0H24}.
However, long-video question answering remains highly challenging because query-relevant evidence is often sparse, brief, and temporally scattered across lengthy video contexts. Critical cues, such as key objects, actions, scene transitions, textual information, and spoken content, may appear only momentarily, while the majority of frames are redundant or irrelevant~\cite{DBLP:conf/cvpr/00010XG25,DBLP:conf/cvpr/IslamNWBT25}. Consequently, VLMs often struggle to effectively model long-range temporal dependencies and accurately identify truly relevant evidence from long-form videos.

\begin{figure}[t]
\includegraphics[width=0.5\textwidth]{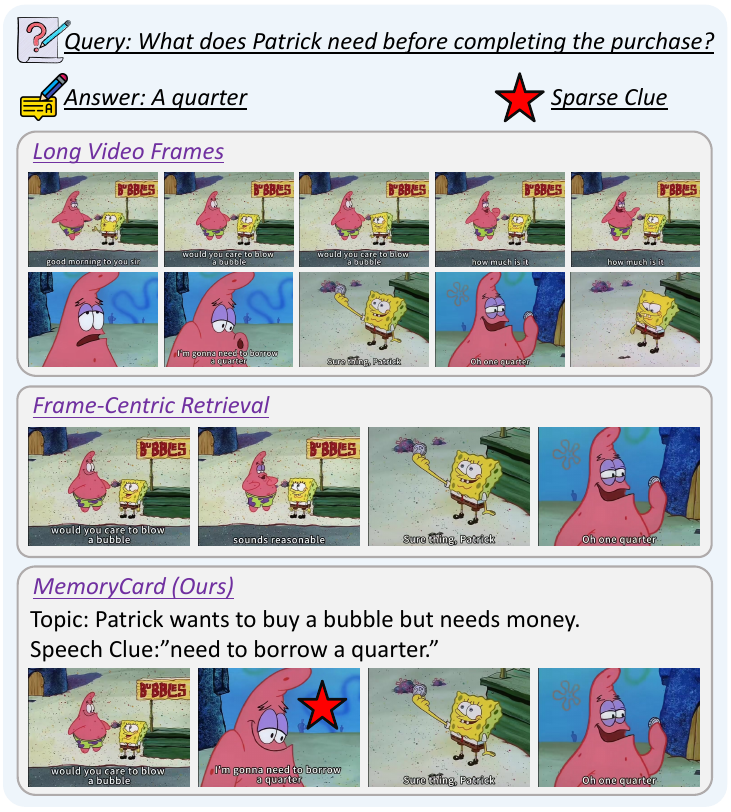}
\caption{A motivational example illustrating the differences between Memory Cards and existing frame-based evidence construction methods.}
\vspace{-16pt}
\label{fig:intro}
\end{figure}

Existing methods primarily address this bottleneck through frame-centric clue filtering and extraction. The most straightforward strategy is to employ uniform frame sampling to compress long-form videos. However, this approach relies on sparsely sampled frames to represent the entire video, which may overlook crucial evidence required for accurate question answering~\cite{DBLP:journals/corr/abs-2307-04192,DBLP:journals/corr/abs-2507-00033}.
To overcome the limitations of uniform sampling, recent approaches aim to better represent long-form videos by selecting query-relevant frames~\cite{DBLP:conf/cvpr/BuchNAS25,DBLP:conf/iclr/YuJWCJZXSZWZS25}, pruning or compressing frame-derived visual tokens~\cite{DBLP:conf/cvpr/TaoQYSW25}, and dynamically adapting input resolutions under fixed computational budgets~\cite{DBLP:conf/acl/HuangZH25}. These methods help preserve informative clues for question answering while reducing redundant visual noise.
Although effective in reducing noisy visual information, these methods still treat raw frames or frame-derived representations as the fundamental units of evidence, which are inherently low in semantic density. Video frames can capture objects, scenes, and instantaneous action states; however, they often provide sparse and fragmented evidence. As a result, they fail to organize continuous frames into semantically coherent events or to associate different events for comprehensive video understanding~\cite{zacks2007event,kurby2008segmentation}. This limitation further hinders VLMs from forming coherent event-level units, capturing continuous video semantics, and performing effective long-range reasoning~\cite{DBLP:conf/emnlp/LiaoEWZZMT24}.

In this work, we propose \method{}, a video-memory-based augmentation framework that organizes videos into \textit{Memory Cards} to facilitate long-video question answering. As illustrated in Figure~\ref{fig:intro}, \method{} goes beyond isolated frame sampling by constructing self-contained Memory Cards with event-level contextual cues. Specifically, \method{} first employs a VLM to segment the video into semantically coherent sessions, each corresponding to a distinct topic or event. 
It then performs intensive reading over each session to generate an event-level video gist, consisting of a VLM-generated topic and aligned utterances, and selects representative visual moments from the same session. To preserve and utilize these multimodal clues, \method{} renders the video gist and representative visual moments into a unified Memory Card. This design transforms sparse frame-level clues into high-density multimodal evidence while remaining compatible with standard image-based VLM pipelines~\cite{visrag}.

Experiments on three long-video question answering benchmarks show that \method{} consistently improves performance under comparable visual budgets. Ablation studies further show that the gains are not merely brought by retrieval, but by the proposed evidence representation: self-read semantic session construction, high-density Memory Card rendering, and temporal clue organization. Additional analyses demonstrate that Memory Cards preserve fine-grained visual details while maintaining event-level temporal context, supporting them as effective multimodal evidence units for offline and reusable long-video understanding.
\section{Related Work}
\label{sec:related_work}

Vision-Language Models (VLMs) have achieved substantial progress in video understanding and temporal reasoning. Existing studies improve long-video understanding primarily from two perspectives. One line of work strengthens the intrinsic video modeling capability of VLMs through temporal-aware modeling, large-scale video instruction tuning, and long-video QA-oriented training~\cite{DBLP:conf/cvpr/RenYL0H24,DBLP:journals/tmlr/ZhangZLZYZWTLL25,DBLP:journals/tmlr/ZhangWLLMLL25}. Another line focuses on extending or compressing long visual contexts via long-context transfer, streaming encoding, and hierarchical compression~\cite{DBLP:conf/eccv/WengHHCZ24,DBLP:conf/nips/QianDZZDLW24,DBLP:journals/tmlr/ZhangZLZYZWTLL25}. Despite these advances, the performance of long-video question answering under constrained visual-token budgets still critically depends on whether compact, relevant, and semantically informative evidence can be effectively accessed from lengthy videos~\cite{luo2026quota, cho2026floc, liu2025keyframe}.

\begin{figure*}[t]
\includegraphics[width=\textwidth]{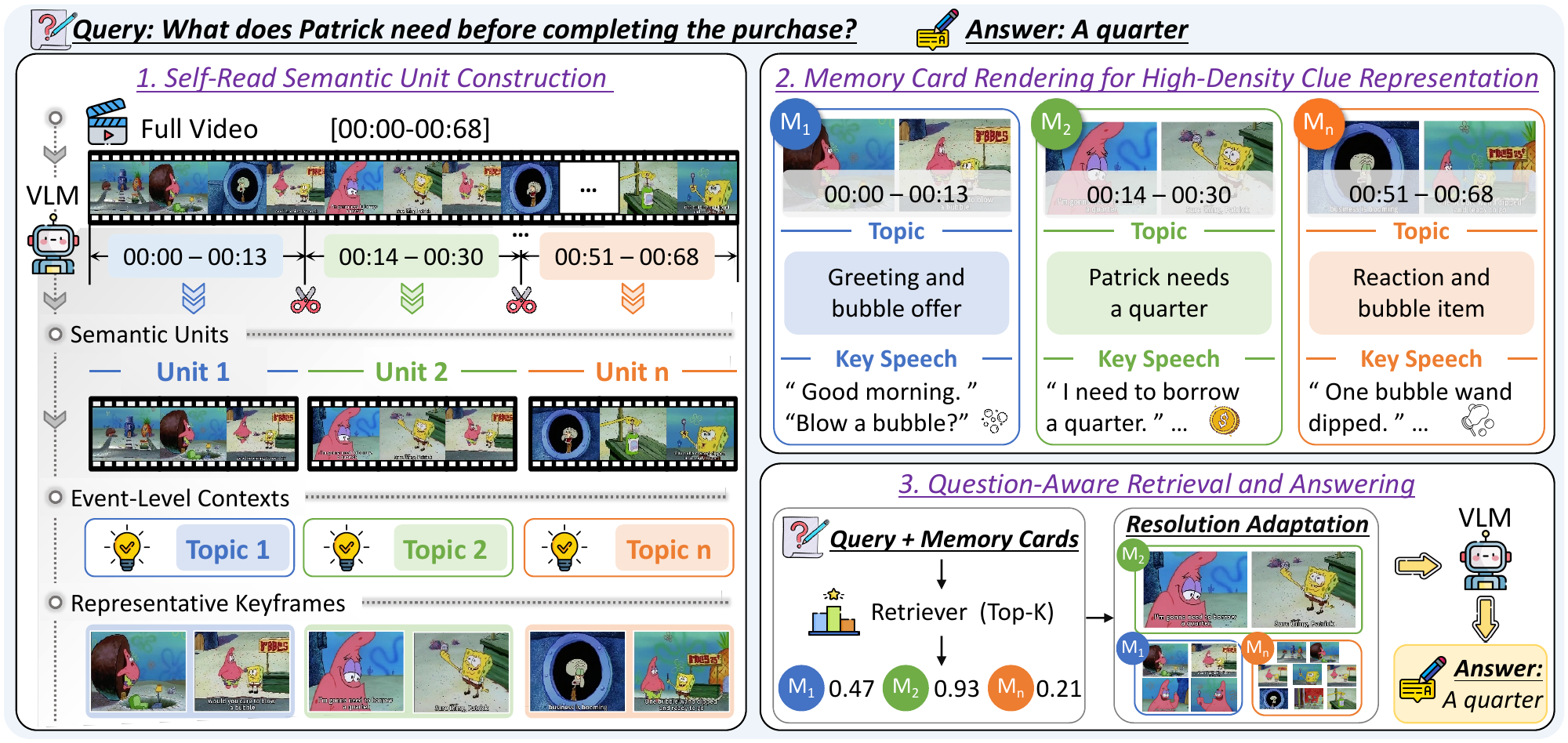}
\caption{
Overview of the \method{} framework. \method{} constructs semantic units by self-reading long-form videos together with aligned utterances, renders representative visual moments and event-level video gists into Memory Cards, and retrieves relevant cards for question answering. The retrieved cards are assigned adaptive input resolutions according to their relevance, reordered based on their original temporal positions, and then fed into the answering VLM.}
\vspace{-10pt}
\label{fig:framework}
\end{figure*}

Recent studies address this evidence bottleneck mainly through frame-centric selection and structured access mechanisms. Some methods identify informative frames or keyframes using VLM-based scoring, adaptive sampling, query-aware retrieval, or generation-time feedback~\cite{DBLP:conf/cvpr/HuGNZTN0SHYC25,DBLP:conf/cvpr/TangQXTJY25,zhang2025qframe,yao-etal-2025-expandr, yao2025generative}. Other approaches organize video access through hierarchical structures, retrieval augmentation, or agent-based multi-step reasoning~\cite{DBLP:conf/cvpr/WangYSYCBB25,DBLP:journals/corr/abs-2411-13093,DBLP:conf/aaai/ChenDX25,DBLP:conf/eccv/WangZZY24}. Recent studies further incorporate complementary textual cues, such as subtitles, transcripts, or OCR information, alongside visual inputs, while training-free approaches improve long-video access through retrieval and memory-based mechanisms~\cite{he2026vsi,dorovatas2026recurrent, di2025streaming}. Although these methods improve long-video access under limited visual-token budgets, their fundamental clue units are still largely restricted to frames, clips, frame-derived representations, tool-specific visual states, or separately provided visual-textual evidence. Such units effectively capture local visual content, yet often lack the event-level semantic context required to interpret sparse visual clues.

External memory modeling provides a promising direction for organizing long video context and has been widely explored in long-context reasoning for text, where memories are used to store interaction histories, user preferences, or factual knowledge beyond limited context windows~\cite{DBLP:conf/emnlp/KangJZB25,DBLP:journals/corr/abs-2510-18866,DBLP:journals/corr/abs-2602-11182}. 
For videos, MovieChat compresses dense frame-level visual tokens into sparse long-term memories to reduce the cost of long-video understanding~\cite{song2024moviechat}, and MovieChat+ further introduces question-aware memory consolidation to preserve more visual tokens from query-relevant segments~\cite{song2025moviechat+}. 
Other memory-based approaches preserve long-range information through fixed-size memory or hierarchical backtracking~\cite{zuo2025videolucy}, and maintain historical visual states, memory tokens, or model-internal memories~\cite{DBLP:conf/cvpr/0004LJJCSSL24,DBLP:conf/acl/WangSZWCN25,DBLP:conf/cvpr/DikoWSSP25}. 
While effective for extending contextual capacity, these methods mainly treat memory as latent visual-token states or model-internal representations, and even question-aware variants use the query to control the compression process rather than constructing a reusable event-level memory bank. 
In contrast, \method{} first self-reads the video and aligned utterances to build question-independent semantic units, and then renders representative visual moments with event-level video gists into image-based Memory Cards. 
These cards serve as explicit, reusable, and multimodal evidence units that are naturally compatible with standard image-based VLM pipelines.
\section{Methodology}
\label{sec:method}

As illustrated in Figure~\ref{fig:framework}, this section introduces \textsc{\method{}}, a video-memory-based augmentation framework that organizes long-form videos into self-contained \textit{Memory Cards} for long-video question answering. We first formulate long-form video QA as a retrieval-based task and highlight the limitations of frame-centric clue units (Sec.~\ref{sec:framework}). We then describe how \method{} constructs a Memory Card bank by segmenting the video into semantic sessions, generating an event-level video gist for each session, and rendering representative visual moments together with the corresponding gist into Memory Cards (Sec.~\ref{sec:memcard}).

\subsection{Preliminaries of Long-Form Video Question Answering}
\label{sec:framework}

Given a long-form video $V$ and a question $q$, the goal of long-form video question answering is to predict the answer $y$ from the long temporal context of the video. We represent the video input as:
\begin{equation}
    V=\{(f_i, t_i)\}_{i=1}^N,
    \label{eq:video_input}
\end{equation}
where $N$ denotes the total number of frames in the video $V$. Based on the video input, we further construct two aligned modalities: the visual observation set $\mathcal{F}=\{f_1,\dots,f_N\}$ and the transcript set $\mathcal{T}=\{t_1,\dots,t_N\}$, where each $t_i$ denotes the transcript aligned with frame $f_i$, such as subtitles or speech transcripts when available.

\textbf{Retrieval-Augmented Video QA.}
Existing retrieval-augmented methods usually follow a retrieve-then-answer pipeline~\cite{DBLP:journals/corr/abs-2312-04931,DBLP:conf/cvpr/MinBNCS24}. Given a question $q$, the retriever selects the top-ranked visual clues from the raw frame pool according to question-frame similarity:
\begin{equation}
    \Tilde{\mathcal{F}}
    =
    \operatorname{Top-}k\big(\mathrm{sim}(q,\mathcal{F})\big),
    \label{eq:retrieve_clue}
\end{equation}
where $\Tilde{\mathcal{F}}$ denotes the retrieved frame-level evidence, $k$ denotes the number of selected visual observations, and $\mathrm{sim}(\cdot,\cdot)$ measures the relevance between the question and the candidate visual observations in $\mathcal{F}$.
The answering VLM then predicts the answer based on the question and the retrieved visual clues:
\begin{equation}
    y=\mathrm{VLM}(q,\Tilde{\mathcal{F}}),
    \label{eq:answer_from_clue}
\end{equation}
where $y$ denotes the predicted answer.

For frame-centric methods, the retrieved clues are selected from the raw-frame pool $\mathcal{F}$~\cite{DBLP:conf/cvpr/GiaLDMNL025}. Although simple and compatible with image-based VLMs, such clue units are semantically sparse and temporally fragmented. A raw frame only records an instantaneous visual state, without explicit event boundaries, event-level context, or aligned spoken content~\cite{DBLP:conf/cvpr/Huang0CS024}. Moreover, retrieving frames independently can break the temporal and semantic continuity of the original video, hindering cross-event association and long-range reasoning~\cite{DBLP:conf/emnlp/LiaoEWZZMT24}.

\textbf{Memory-Based Clue Construction.}
The key limitation of frame-centric retrieval lies in the clue unit itself: an isolated frame provides only sparse visual evidence and lacks the event-level context needed for long-video reasoning. To address this limitation, \method{} constructs a Memory Card bank from the original video:
\begin{equation}
    \mathcal{M}
    =
    \mathrm{MemConstruct}(V),
    \label{eq:memory_bank_overview}
\end{equation}
where $\mathcal{M}$ denotes the constructed Memory Card bank and $\mathrm{MemConstruct}(\cdot)$ is the memory construction method in Sec.~\ref{sec:memcard}.  
Specifically, the memory set $\mathcal{M}$ consists of multiple Memory Cards:
\begin{equation}
    \mathcal{M}=\{M_i\}_{i=1}^{J},
    \label{eq:memory_bank_composition}
\end{equation}
where $J$ denotes the number of semantic sessions, and each $M_i$ is an image-based memory unit derived from one semantic session. 
Each Memory Card anchors representative visual moments and enriches them with an event-level video gist. 
Through this design, \method{} converts sparse frame-level observations into compact multimodal evidence units.

Once constructed, $\mathcal{M}$ serves as the retrieval pool for \method{}. Given a question $q$, \method{} retrieves the most relevant Memory Cards according to the question-card similarity:
\begin{equation}
    \Tilde{\mathcal{M}}
    =
    \operatorname{Top-}k \big(\mathrm{sim}(q,\mathcal{M})\big),
    \label{eq:memory_card_retrieval}
\end{equation}
where $M_i$ denotes a Memory Card in $\mathcal{M}$, $\mathrm{sim}(\cdot,\cdot)$ measures question-card relevance, $K_{\mathrm{ret}}$ denotes the number of retrieved Memory Cards, and $\Tilde{\mathcal{M}}$ denotes the retrieved card set.

The answering VLM then predicts the answer based on the question and the retrieved event-level Memory Card clues:
\begin{equation}
    y=\mathrm{VLM}(q,\Tilde{\mathcal{M}}),
    \label{eq:answer_from_memory_cards}
\end{equation}
where $y$ denotes the predicted answer generated from the retrieved Memory Card evidence $\Tilde{\mathcal{M}}$.

The retrieved Memory Cards are assigned input resolutions according to their relevance and reordered by their original temporal positions before being fed into the answering VLM.

\subsection{Event-Level Multimodal Evidence Construction}
\label{sec:memcard}
This subsection instantiates $\mathrm{MemConstruct}(V)$ by converting the input video into an event-level Memory Card bank. The construction process consists of two stages: adaptive video segmentation and Memory Card construction.

\textbf{Adaptive Video Segmentation.}
To construct event-level evidence, \method{} first prompts a VLM to analyze the original video input $V$. The VLM then partitions the video into $J$ semantic sessions according to event, topic, or scene transitions:
\begin{equation}
    \{(b_i,e_i)\}_{i=1}^{J}
    = \mathrm{VLM}(\text{Instruct}_\text{seg}, V),
    \label{eq:semantic_session_segmentation}
\end{equation}
where $\text{Instruct}_\text{seg}$ indicates the instruction for session segmentation. $s_i$ denotes the $i$-th semantic session identified by the VLM, while $b_i$ and $e_i$ denote the corresponding start and end timestamps, respectively. Based on these boundaries, each semantic session $s_i$ is extracted from the original video $V$ as:
\begin{equation}
    s_i = V[b_i: e_i].
\end{equation}

\textbf{Memory Card Construction.}
Given a semantic session $s_i = V[b_i:e_i]$, \method{} first constructs an event-level video gist $g_i$ by summarizing the video segment $s_i$, and then selects representative frames to characterize the session content.

Specifically, the model summarizes the visual and textual information within the session, i.e., $\mathcal{F}_{b_i:e_i}$ and $\mathcal{T}_{b_i:e_i}$, to produce an event-level summary of both language and vision:
\begin{equation}
g_i, I_i = \mathrm{VLM} (\text{Instruct}_\text{gist}, \mathcal{F}_{b_i:e_i}, \mathcal{T}_{b_i:e_i}),
\end{equation}
where $\text{Instruct}_\text{gist}$ denotes the instruction that prompts the model to extract the key clues from session $s_i$ to form the corresponding gist $g_i$, and to generate the indices $I_i$ of representative frames that best characterize the session.
Next, \method{} selects representative visual frames from the frame set $\mathcal{F}_{b_i:e_i}$ according to the generated frame indices $I_i$ to form the representative frame set $\tilde{\mathcal{F}}_{b_i:e_i}$:
\begin{equation}
\tilde{\mathcal{F}}_{b_i:e_i}
=
\mathrm{Select}(\mathcal{F}_{b_i:e_i}, I_i),
\label{eq:representative_frame_selection}
\end{equation}
where $\tilde{\mathcal{F}}_{b_i:e_i} \subseteq \mathcal{F}_{b_i:e_i}$ denotes the representative visual moments selected from session $s_i$. Since the selection process is guided by the event-level video gist, the resulting visual moments are both visually informative and semantically aligned with the session content.
Finally, each Memory Card is constructed by rendering the selected visual moments, the event-level video gist, and the session boundary information into a unified image-based clue representation:
\begin{equation}
M_i
=
\mathrm{Render}\big(\tilde{\mathcal{F}}_{b_i:e_i}, g_i, (b_i,e_i)\big),
\label{eq:single_memory_card}
\end{equation}
where $M_i$ denotes the Memory Card representing the $i$-th semantic session $s_i$.

\section{Experimental Methodology}
\label{sec:experiments}

This section describes the experimental setup for evaluating \method{} on long-video question answering benchmarks. 
We use \method{} as a test-time augmentation framework: the answering VLMs are kept fixed, and no model parameters are updated. 
To ensure a fair evaluation and avoid information leakage, all Memory Cards are constructed prior to retrieval in a question-agnostic manner, without access to downstream questions, answer options, or ground-truth answers.

\textbf{Datasets.}
We evaluate \method{} on three long-video question answering benchmarks: Video-MME~\cite{DBLP:conf/cvpr/FuDLLRZWZSZCLLZ25}, MLVU~\cite{zhou2024mlvu}, and LongVideoBench~\cite{DBLP:conf/nips/WuLCL24}. 
All benchmarks follow the multiple-choice protocol, and accuracy is used as the evaluation metric. 
For Video-MME, we report overall accuracy and duration-wise accuracy on short, medium, and long videos. 
For MLVU and LongVideoBench, we report overall accuracy on the evaluated split. 
Additional dataset details and evaluation protocol are provided in Appendix~\ref{app:dataset_protocol}.

\textbf{Baselines.}
We compare \method{} with representative video VLMs and efficient long-video understanding methods. 
These methods cover video-language models, long-context or compression-based video models, frame-selection methods, and memory-based long-video approaches. 
Brief descriptions of the compared methods are provided in Appendix~\ref{app:compared_methods}. 
For controlled comparisons, we use three answering backbones: Qwen2-VL-7B, Qwen3-VL-8B, and MiniCPM-V-4.5. 
For each backbone, we compare the original model with its \method{}-augmented variant while keeping the answering model, prompt format, decoding configuration, answer extraction rule, and evaluation protocol unchanged.

\begin{table*}[t]
\centering
\small
\setlength{\tabcolsep}{4.2pt}
\renewcommand{\arraystretch}{1.08}
\begin{tabular}{@{}lcccccccc@{}}
\toprule
\multirow{3}{*}{\textbf{Method}} 
& \multirow{3}{*}{\textbf{LLM Size}} 
& \multirow{3}{*}{\textbf{\#Frames}} 
& \multicolumn{4}{c}{\textbf{Video-MME}}
& \multirow{3}{*}{\textbf{MLVU}} 
& \multirow{3}{*}{\textbf{LongVideoBench}} \\
\cmidrule(lr){4-7}

& & 
& \textbf{Overall}
& \textbf{Short}
& \textbf{Medium}
& \textbf{Long}
& & \\

\multicolumn{3}{c}{\textit{Avg. Video Duration}}
& \textit{17min}
& \textit{1.3min}
& \textit{9min}
& \textit{41min}
& \textit{12min}
& \textit{12min} \\
\midrule

Video-LLaVA        & 7B & 8       & 39.9 & 45.3 & 38.0 & 36.2 & 47.3 & 39.1 \\
Qwen-VL            & 7B & 8       & 41.1 & 46.9 & 38.7 & 37.8 & -    & -    \\
VideoChat2         & 7B & 8       & 39.5 & 48.3 & 37.0 & 33.2 & 44.5 & 39.3 \\
ShareGPT4Video     & 8B & 16      & 39.9 & 48.3 & 36.3 & 35.0 & 46.4 & 41.8 \\
MovieChat          & 7B & *       & 38.2 & -    & -    & 33.4 & 25.8 & -    \\ 
MovieChat+         & 7B & *       & 44.5 & 49.3 & 44.5 & 39.7 & 31.2 & 40.4\\
Video-XL           & 7B & 128/256 & 55.5 & 64.0 & 53.2 & 49.2 & 64.9 & -    \\
LLaVA-OneVision    & 7B & *       & 58.2 & -    & -    & -    & 64.7 & 56.3 \\
Video-CCAM         & 9B & 96      & 50.3 & 61.9 & 49.2 & 39.6 & 58.5 & -    \\
Frame-Voyager      & 8B & 8       & 57.5 & 67.3 & 56.3 & 48.9 & 65.6 & -    \\
LongVU             & 7B & *       & 60.9 & 64.7 & 58.2 & 59.5 & 65.4 & -    \\
\hdashline

Qwen2-VL-Video     & 7B & 8       & 53.0 & 64.1 & 49.3 & 45.6 & 55.6 & 51.0 \\
Qwen2-VL           & 7B & 8       & 53.7 & 65.0 & 50.7 & 45.3 & 56.9 & 53.5 \\
\rowcolor{memoryblue} \quad \textbf{+MemoryCard}  & 7B & \MCFrames
& \textbf{60.5} & \textbf{69.7} & \textbf{61.0} & \textbf{50.9} 
& \textbf{65.7} & \textbf{58.4} \\
\hdashline

Qwen3-VL-Video     & 8B & 8       & 56.6 & 67.7 & 51.7 & 50.3 & 56.9 & 54.8 \\
Qwen3-VL           & 8B & 8       & 57.4 & 68.8 & 53.1 & 50.2 & 57.2 & 56.3 \\
\rowcolor{memoryblue}\quad \textbf{+MemoryCard}  & 8B & \MCFrames
& \textbf{64.7} & \textbf{72.2} & \textbf{64.8} & \textbf{54.7} 
& \textbf{66.5} & \textbf{60.1} \\
\hdashline

MiniCPM-V-4.5-Video & 8B & 8       & 59.7 & 68.4 & 58.9 & 51.8 & 58.9 & 56.3 \\
MiniCPM-V-4.5       & 8B & 8       & 59.9 & 69.6 & 58.2 & 52.0 & 57.0 & 55.6 \\
\rowcolor{memoryblue}\quad \textbf{+MemoryCard}   & 8B & \MCFrames
& \textbf{67.2} & \textbf{76.0} & \textbf{67.6} & \textbf{58.0} 
& \textbf{69.4} & \textbf{62.0} \\
\bottomrule
\end{tabular}
\caption{Overall Performance of \method{}. Comparison with representative VLMs on Video-MME, MLVU, and LongVideoBench.
All results are reported as accuracy.
For Video-MME, we report the overall accuracy and the results on short, medium, and long videos.
For \method{}, 
4,
\textcolor{midgray}{8}, and
\textcolor{lightgray}{32}
denote the numbers of memory-card inputs at high, medium, and low resolutions, respectively.
The lighter the color, the lower the frame resolution.
The best results are shown in \textbf{bold}.
}
\vspace{-6pt}
\label{tab:main_results}
\end{table*}

\textbf{Implementation Details.}
For each video, \method{} first constructs a reusable, question-agnostic Memory Card bank, and then performs question-aware retrieval at test time.
We use Qwen3-ASR~\cite{DBLP:journals/corr/abs-2601-21337} for timestamped speech transcription, Qwen3-VL-8B~\cite{bai2025qwen3vltechnicalreport} as the self-read VLM, and LongCLIP~\cite{DBLP:conf/eccv/ZhangZDZW24} as the CLIP-style retriever.
Unless otherwise specified, we retrieve $k_{\mathrm{ret}}=44$ Memory Cards, with $4$, $8$, and $32$ cards assigned to high, medium, and low input resolutions, respectively.
All evaluations are conducted with the lmms-eval framework~\cite{DBLP:conf/naacl/ZhangLZPCHLZYLL25}.
More implementation details are provided in Appendix~\ref{app:implementation_details}.
We additionally report the computational cost of \method{}, including both offline Memory Card construction and online per-question inference, in Appendix~\ref{sec:runtime_analysis}.
\section{Evaluation Results}
\label{sec:result}

We evaluate \method{} from four perspectives: overall performance, rendered card components, semantic session construction, and the use of Memory Cards under constrained visual budgets. 
Together, these analyses examine whether event-level multimodal evidence units provide more effective inputs for long-video question answering than frame-centric clues. 

\subsection{Overall Performance}
\label{sec:main_results}
As shown in Table~\ref{tab:main_results}, the main results demonstrate that \method{} provides a consistently stronger evidence interface for long-video question answering.
Across different benchmarks and answering backbones, the original VLMs obtain clear gains after being augmented with Memory Cards.
Since the answering model, prompt format, decoding configuration, answer extraction rule, and evaluation protocol are kept unchanged, these improvements mainly come from the way video evidence is constructed and supplied to the VLM, rather than from model retraining or a stronger inference setting.

This result directly supports the motivation of \method{}.
Long-video QA is often limited not only by the amount of visual input, but also by the quality of the evidence units given to the model.
Raw frames are sparse and fragmented: they may capture objects or instantaneous actions, but they usually do not provide explicit event boundaries, spoken clues, or temporal context.
Memory Cards address this limitation by packaging representative visual moments together with event-level semantic gists, aligned speech clues, and temporal metadata.
As a result, the retrieved inputs become more self-contained and semantically informative, making them easier for the answering VLM to interpret.
\begin{table*}[t]
\centering
\small
\setlength{\tabcolsep}{8.0pt}
\renewcommand{\arraystretch}{1.08}
\begin{tabular}{lcccccc}
\toprule
\multirow{2}{*}{\textbf{Method}} 
& \multicolumn{4}{c}{\textbf{Video-MME}} 
& \multirow{2}{*}{\textbf{MLVU}} 
& \multirow{2}{*}{\textbf{LongVideoBench}} \\
\cmidrule(lr){2-5}
& \textbf{Overall} 
& \textbf{Short} 
& \textbf{Medium} 
& \textbf{Long} 
& 
& \\
\midrule

Qwen2-VL & 53.7 & 65.0 & 50.7 & 45.3 & 56.9 & 53.5 \\

\quad + Raw-frame Retrieval & 57.3 & 69.0 & 55.7 & 47.3 & 64.6 & 56.3 \\

\rowcolor{memoryblue}
\textbf{MemoryCard} 
& \textbf{60.5} & \textbf{69.7} & \textbf{61.0} & \textbf{50.9} & \textbf{65.7} & \textbf{58.4} \\

\quad w/o Speech Transcript & 57.8 & 68.9 & 55.6 & 48.8 & 65.1 & 56.6 \\

\quad w/o Topic & 58.6 & 68.0 & 59.1 & 48.6 & 64.7 & 56.7 \\

\quad w/o Temporal Span & 59.5 & 69.1 & 59.7 & 49.6 & 64.9 & 57.2 \\

\midrule

Qwen3-VL & 57.4 & 68.8 & 53.1 & 50.2 & 57.2 & 56.3 \\

\quad + Raw-frame Retrieval & 60.8 & 69.8 & 58.9 & 53.6 & 64.9 & 57.2 \\

\rowcolor{memoryblue}
\textbf{MemoryCard} 
& \textbf{64.7} & \textbf{72.2} & \textbf{64.8} & \textbf{54.7} & \textbf{66.5} & \textbf{60.1} \\

\quad w/o Speech Transcript & 61.3 & 70.6 & 60.8 & 52.4 & 65.9 & 58.8 \\

\quad w/o Topic & 62.7 & 70.2 & 63.9 & 54.0 & 64.7 & 58.7 \\

\quad w/o Temporal Span & 63.1 & 71.3 & 64.1 & 53.9 & 65.8 & 59.1 \\

\midrule

MiniCPM-V-4.5 & 59.9 & 69.6 & 58.2 & 52.0 & 57.0 & 55.6 \\

\quad + Raw-frame Retrieval & 65.1 & 74.6 & 65.3 & 55.3 & 67.1 & 60.8 \\

\rowcolor{memoryblue}
\textbf{MemoryCard} 
& \textbf{67.2} & \textbf{76.0} & \textbf{67.6} & \textbf{58.0} & \textbf{69.4} & \textbf{62.0} \\ 

\quad w/o Speech Transcript & 65.0 & 74.2 & 64.3 & 56.6 & 68.5 & 61.0 \\

\quad w/o Topic & 66.6 & 75.1 & 66.9 & 57.8 & 68.4 & 61.3 \\

\quad w/o Temporal Span & 66.8 & 75.8 & 66.7 & 57.9 & 67.6 & 60.9 \\

\bottomrule
\end{tabular}
\caption{Ablation study of \method{} components and raw-frame retrieval baseline on three video question answering benchmarks with different answering backbones. The best results within each answering backbone are highlighted in \textbf{bold}.}
\vspace{-10pt}
\label{tab:ablation_components}
\end{table*}

The consistent improvements across Qwen2-VL, Qwen3-VL, and MiniCPM-V-4.5, as well as across different video durations in Video-MME, show that the benefit of \method{} is not tied to a specific backbone or temporal scale.
Different VLMs can all benefit from receiving structured event-level evidence, and Memory Cards remain effective for both local clues in shorter videos and temporally dispersed clues in longer videos.
This suggests that \method{} improves the input representation itself by converting video content into compact, coherent, and retrieval-friendly multimodal evidence under comparable visual-token budgets.

\subsection{Ablation Study}
\label{subsec:ablation}

\begin{table}[t]
\centering
\footnotesize
\setlength{\tabcolsep}{3pt}
\renewcommand{\arraystretch}{1.12}
\begin{tabular}{@{}lccc@{}}
\toprule
\textbf{Method} 
& \makecell{\textbf{Video-MME}\\\textbf{Overall}} 
& \textbf{MLVU} 
& \makecell{\textbf{LongVideo}\\\textbf{Bench}} \\
\midrule

\textbf{MemoryCard} 
& \textbf{64.7} & \textbf{66.5} & 60.1 \\

w/ Uniform-frame Units 
& 62.5 & 65.3 & 59.1 \\

w/ Fixed-length Units 
& 63.1 & 66.2 & 58.8 \\

w/ Shot-based Units 
& 63.4 & 65.9 & \textbf{61.1} \\
\bottomrule
\end{tabular}
\caption{Effect of different memory unit construction strategies with Qwen3-VL as the answering backbone. For Video-MME, we report the overall accuracy. More comprehensive results are reported in Appendix Table~\ref{tab:ablation_units}.}
\vspace{-12pt}
\label{tab:ablation_units_qwen3}
\end{table}

\begin{figure}[t]
    \centering
    \resizebox{0.85\columnwidth}{!}{
        \input{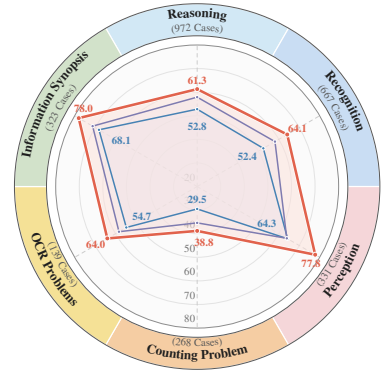}
    }
    \caption{Category-Wise Accuracies (\%) of Qwen3-VL-Video, Qwen3-VL, and \method{} on six task categories in Video-MME.}
    \vspace{-14pt}
    \label{fig:radar_qwen3_videomme}
\end{figure}
Table~\ref{tab:ablation_components} evaluates the contribution of the rendered event-level video gist, speech transcript, topic description, and temporal metadata. 
Raw-frame retrieval serves as a direct control because it uses question-aware retrieval while keeping the evidence unit as raw frames. 
The gap between raw-frame retrieval and the full \method{} shows that selecting relevant frames is not sufficient when the retrieved clues remain semantically sparse and temporally isolated.

Removing individual fields further confirms the role of multimodal card rendering. 
Speech Transcript refers to the aligned utterance text rendered in each Memory Card, and its removal tests the contribution of spoken information. 
Topic provides the unit-level semantic summary, while Temporal Span grounds each card within the original video structure. 
These components complement the visual gist by binding visual content with semantic, speech, and temporal information. 
Thus, the effectiveness of \method{} comes not only from retrieving relevant evidence, but also from rendering the evidence into a self-contained and interpretable Memory Card. 
Detailed variant definitions are provided in Appendix~\ref{app:component_ablation_settings}.

\subsection{Effectiveness of Semantic Memory Sessions}
\label{subsec:semantic_memory_units}
Table~\ref{tab:ablation_units_qwen3} studies how memory sessions should be formed before rendering. 
Adaptive semantic sessions achieve the strongest overall results on Video-MME and MLVU, while remaining competitive on LongVideoBench. This indicates that the gain of \method{} is not obtained by adding text around sampled frames; the visual anchor and event-level context should describe the same event.


Uniform or fixed-length sessions may split related evidence or merge unrelated events, while shot-based sessions mainly capture visual transitions rather than semantic boundaries. 
In contrast, VLM self-reading constructs content-aware sessions, making the rendered Memory Cards more coherent for retrieval and answering.
This supports the central design of \method{}: Memory Cards are effective because they convert long videos into semantically coherent event-level evidence units rather than isolated frame-level clues. 
Complete results are provided in Appendix~\ref{app:unit_construction_analysis}.

Figure~\ref{fig:radar_qwen3_videomme} further analyzes where the overall gains come from. 
\method{} improves multiple task categories rather than only a single capability such as OCR or object recognition, suggesting that Memory Cards provide a generally useful evidence representation for different types of challenging long-video questions.

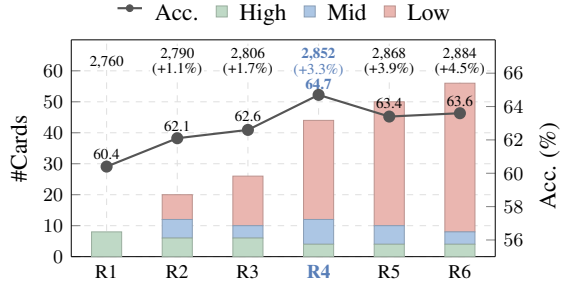
\begin{figure}[t]
    \centering
    \resizebox{\columnwidth}{!}{
        \begin{tikzpicture}

\begin{axis}[
    name=cardaxis,
    width=0.90\linewidth,
    height=0.46\linewidth,
    scale only axis,
    ybar stacked,
    bar width=14pt,
    ymin=0,
    ymax=70,
    xmin=0.5,
    xmax=6.5,
    enlarge x limits=false,
    ylabel={\#Cards},
    ylabel style={font=\normalsize},
    xtick={1,2,3,4,5,6},
    xticklabels={
        R1,
        R2,
        R3,
        \textcolor{accentcol}{\textbf{R4}},
        R5,
        R6
    },
    xticklabel style={font=\small},
    ytick={0,10,20,30,40,50,60},
    yticklabel style={font=\small},
    grid=major,
    grid style={dashed,gray!25},
    axis y line*=left,
    axis x line*=bottom,
    clip=false,
]

\addplot+[
    fill=highcol,
    draw=highcol!80!black
] coordinates {
    (1,8) (2,6) (3,6) (4,4) (5,4) (6,4)
};

\addplot+[
    fill=midcol,
    draw=midcol!80!black
] coordinates {
    (1,0) (2,6) (3,4) (4,8) (5,6) (6,4)
};

\addplot+[
    fill=lowcol,
    draw=lowcol!80!black
] coordinates {
    (1,0) (2,8) (3,16) (4,32) (5,40) (6,48)
};

\node[font=\scriptsize] at (axis cs:1,4) {8};

\node[font=\scriptsize] at (axis cs:2,3) {6};
\node[font=\scriptsize] at (axis cs:2,9) {6};
\node[font=\scriptsize] at (axis cs:2,16) {8};

\node[font=\scriptsize] at (axis cs:3,3) {6};
\node[font=\scriptsize] at (axis cs:3,8) {4};
\node[font=\scriptsize] at (axis cs:3,18) {16};

\node[font=\scriptsize] at (axis cs:4,2) {4};
\node[font=\scriptsize] at (axis cs:4,8) {8};
\node[font=\scriptsize] at (axis cs:4,28) {32};

\node[font=\scriptsize] at (axis cs:5,2) {4};
\node[font=\scriptsize] at (axis cs:5,7) {6};
\node[font=\scriptsize] at (axis cs:5,30) {40};

\node[font=\scriptsize] at (axis cs:6,2) {4};
\node[font=\scriptsize] at (axis cs:6,6) {4};
\node[font=\scriptsize] at (axis cs:6,32) {48};

\node[font=\scriptsize,align=center,fill=white,inner sep=0.8pt]
    at (axis cs:1,62.8) {2,760};

\node[font=\scriptsize,align=center,fill=white,inner sep=0.8pt]
    at (axis cs:2,62.8) {2,790\\[-1pt]\scriptsize(+1.1\%)};

\node[font=\scriptsize,align=center,fill=white,inner sep=0.8pt]
    at (axis cs:3,62.8) {2,806\\[-1pt]\scriptsize(+1.7\%)};

\node[
    font=\scriptsize,
    align=center,
    fill=white,
    inner sep=0.8pt,
    text=accentcol
] at (axis cs:4,62.8) {\textbf{2,852}\\[-1pt]\scriptsize(+3.3\%)};

\node[font=\scriptsize,align=center,fill=white,inner sep=0.8pt]
    at (axis cs:5,62.8) {2,868\\[-1pt]\scriptsize(+3.9\%)};

\node[font=\scriptsize,align=center,fill=white,inner sep=0.8pt]
    at (axis cs:6,62.8) {2,884\\[-1pt]\scriptsize(+4.5\%)};

\draw[
    black,
    line width=0.45pt
] (axis cs:0.5,70) -- (axis cs:6.5,70);

\end{axis}

\begin{axis}[
    at={(cardaxis.south west)},
    anchor=south west,
    width=0.90\linewidth,
    height=0.46\linewidth,
    scale only axis,
    ymin=55,
    ymax=68,
    xmin=0.5,
    xmax=6.5,
    enlarge x limits=false,
    axis y line*=right,
    axis x line=none,
    ylabel={Acc. (\%)},
    ylabel style={font=\normalsize, xshift=-8pt},
    ytick={56,58,60,62,64,66},
    yticklabel style={font=\small},
    xtick=\empty,
    clip=false,
]

\addplot[
    linecol,
    line width=1.0pt,
    mark=*,
    mark size=2.3pt,
    mark options={draw=linecol, fill=linecol}
] coordinates {
    (1,60.4)
    (2,62.1)
    (3,62.6)
    (4,64.7)
    (5,63.4)
    (6,63.6)
};

\node[font=\scriptsize] at (axis cs:1,61.15) {60.4};
\node[font=\scriptsize] at (axis cs:2,62.85) {62.1};
\node[font=\scriptsize] at (axis cs:3,63.45) {62.6};
\node[font=\scriptsize\bfseries,text=accentcol] at (axis cs:4,65.35) {64.7};
\node[font=\scriptsize] at (axis cs:5,64.2) {63.4};
\node[font=\scriptsize] at (axis cs:6,64.35) {63.6};

\end{axis}

\node[anchor=south, inner sep=0pt] at ([yshift=0.18cm]cardaxis.north) {%
\normalsize
\begin{tabular}{@{}c@{\hspace{8pt}}c@{\hspace{8pt}}c@{\hspace{8pt}}c@{}}
\tikz[baseline=-0.5ex]{
    \draw[linecol,line width=1.0pt] (0,0.10) -- (0.40,0.10);
    \filldraw[draw=linecol,fill=linecol] (0.20,0.10) circle (0.055);
}
~Acc.
&
\tikz[baseline=-0.25ex]\filldraw[fill=highcol,draw=highcol!80!black] (0,0) rectangle (0.18,0.18);
~High
&
\tikz[baseline=-0.25ex]\filldraw[fill=midcol,draw=midcol!80!black] (0,0) rectangle (0.18,0.18);
~Mid
&
\tikz[baseline=-0.25ex]\filldraw[fill=lowcol,draw=lowcol!80!black] (0,0) rectangle (0.18,0.18);
~Low
\end{tabular}
};

\end{tikzpicture}
    }
    \vspace{-15pt}
    \caption{Resolution Allocation under Comparable Visual Budgets. R1--R6 are detailed in Appendix Table~\ref{tab:4_resolution_allocation_analysis}; bars show card resolutions and the line shows Video-MME accuracy with Qwen3-VL.}
    \label{fig:resolution_allocation}
    \vspace{-12pt}
\end{figure}
The gains are especially relevant for tasks that require local visual details to be interpreted with broader event context, such as perception, recognition, OCR, and information synopsis. 
Counting and reasoning remain more challenging, suggesting that some questions may require denser temporal coverage or stronger aggregation across multiple retrieved cards. 
Complete subtask results are provided in Appendix~\ref{app:subtask_analysis}.

\subsection{Effectiveness of Memory Card Selection and Allocation}
\label{subsec:selection_allocation}
This subsection studies how constructed Memory Cards should be used under a fixed visual-token budget.
After the Memory Card bank is built, \method{} still needs to decide which cards to select, how much visual resolution to assign to each card, and how to organize the selected cards before answer generation.
These steps determine whether the compact event-level evidence can be effectively consumed by the answering VLM.


\textbf{Resolution Allocation.}
Figure~\ref{fig:resolution_allocation} compares different ways of allocating the visual budget among retrieved Memory Cards. 
High-resolution cards preserve fine-grained visual details but reduce event coverage, while low-resolution cards increase coverage but weaken detailed perception. 
The default relevance-aware allocation balances this trade-off by assigning higher resolution to likely answer-critical cards and lower resolution to contextual cards. 
This allows \method{} to preserve both key visual details and broader temporal context within a comparable visual budget. 
Full results are provided in Appendix~\ref{app:resolution_allocation_analysis}.

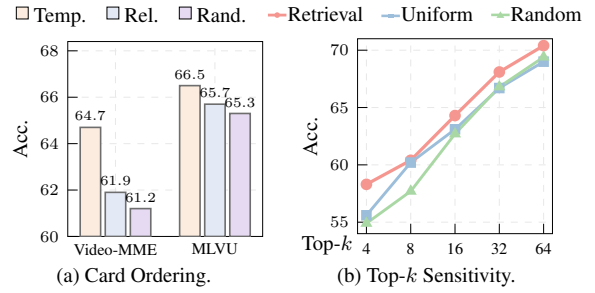
\begin{figure}[t]
    \centering
    \centering

\definecolor{barteal}{RGB}{252,236,223} 
\definecolor{barlav}{RGB}{227,233,245}  
\definecolor{bargrayblue}{RGB}{230,217,242} 

\definecolor{lineret}{RGB}{245,151,144}
\definecolor{lineuni}{RGB}{156,190,219}
\definecolor{lineran}{RGB}{169,216,165}

\definecolor{barborder}{RGB}{110,110,110}

\def\subfigsize{2.95cm}

\captionsetup[subfigure]{font=footnotesize, labelfont=normalfont, skip=1pt}

\resizebox{\columnwidth}{!}{%
\begin{tabular}{@{}c@{\hspace{0.4mm}}c@{}}

\subcaptionbox{Card Ordering.\label{fig:ordering_topk_a}}{%
\begin{tabular}[t]{@{}c@{}}
{\footnotesize
\raisebox{0.25ex}{\tikz{\filldraw[draw=black, fill=barteal] (0,0) rectangle (0.18,0.18);}}~Temp.\hspace{0.65em}
\raisebox{0.25ex}{\tikz{\filldraw[draw=black, fill=barlav] (0,0) rectangle (0.18,0.18);}}~Rel.\hspace{0.65em}
\raisebox{0.25ex}{\tikz{\filldraw[draw=black, fill=bargrayblue] (0,0) rectangle (0.18,0.18);}}~Rand.
}
\\[-7pt]
\begin{tikzpicture}[baseline=(current bounding box.north)]
    \begin{axis}[
        width=\subfigsize,
        height=\subfigsize,
        scale only axis,
        ybar,
        bar width=8.8pt,
        ymin=60,
        ymax=68.4,
        ylabel={Acc.},
        ylabel style={font=\footnotesize, yshift=-2pt},
        ytick={60,62,64,66,68},
        yticklabel style={font=\scriptsize},
        symbolic x coords={Video-MME,MLVU},
        xtick=data,
        xticklabel style={font=\scriptsize, yshift=5pt},
        xtick style={draw=none},
        enlarge x limits=0.48,
        nodes near coords,
        every node near coord/.append style={font=\tiny, yshift=-1pt},
        grid=major,
        grid style={dashed, gray!18},
        major grid style={gray!18},
        tick style={black},
        axis line style={black},
        tick pos=left,
    ]

    \addplot[
        draw=barborder,
        line width=0.7pt,
        fill=barteal
    ] coordinates {
        (Video-MME,64.7)
        (MLVU,66.5)
    };

    \addplot[
        draw=barborder,
        line width=0.7pt,
        fill=barlav
    ] coordinates {
        (Video-MME,61.9)
        (MLVU,65.7)
    };

    \addplot[
        draw=barborder,
        line width=0.7pt,
        fill=bargrayblue
    ] coordinates {
        (Video-MME,61.2)
        (MLVU,65.3)
    };

    \end{axis}
\end{tikzpicture}
\end{tabular}%
}
&
\subcaptionbox{Top-$k$ Sensitivity.\label{fig:ordering_topk_b}}{%
\begin{tabular}[t]{@{}c@{}}
{\footnotesize
\raisebox{0.35ex}{%
\begin{tikzpicture}
    \draw[lineret, line width=1.20pt] (0,0) -- (0.28,0);
    \fill[lineret] (0.14,0) circle (1.20pt);
\end{tikzpicture}}%
Retrieval\hspace{0.55em}
\raisebox{0.35ex}{%
\begin{tikzpicture}
    \draw[lineuni, line width=1.10pt] (0,0) -- (0.28,0);
    \fill[lineuni] (0.10,-0.03) rectangle (0.18,0.03);
\end{tikzpicture}}%
Uniform\hspace{0.55em}
\raisebox{0.35ex}{%
\begin{tikzpicture}
    \draw[lineran, line width=1.10pt] (0,0) -- (0.28,0);
    \fill[lineran] (0.14,0.055) -- (0.08,-0.045) -- (0.20,-0.045) -- cycle;
\end{tikzpicture}}%
Random
}
\\[-7pt]
\begin{tikzpicture}[baseline=(current bounding box.north)]
    \begin{axis}[
        width=\subfigsize,
        height=\subfigsize,
        scale only axis,
        xmin=0.8,
        xmax=5.2,
        ymin=54,
        ymax=71,
        clip=false,
        ylabel={Acc.},
        ylabel style={font=\footnotesize, yshift=-2pt},
        ytick={55,60,65,70},
        yticklabel style={font=\scriptsize},
        xtick={1,2,3,4,5},
        xticklabels={4,8,16,32,64},
        xticklabel style={font=\scriptsize},
        grid=major,
        grid style={dashed, gray!18},
        major grid style={gray!18},
        tick style={black},
        axis line style={black},
        tick pos=left,
    ]

    \addplot[
        color=lineret,
        mark=*,
        mark size=1.80pt,
        line width=1.40pt,
        mark options={fill=lineret, draw=lineret}
    ] coordinates {
        (1,58.3)
        (2,60.4)
        (3,64.3)
        (4,68.1)
        (5,70.4)
    };

    \addplot[
        color=lineuni,
        mark=square*,
        mark size=1.65pt,
        line width=1.25pt,
        mark options={fill=lineuni, draw=lineuni}
    ] coordinates {
        (1,55.6)
        (2,60.2)
        (3,63.1)
        (4,66.7)
        (5,69.0)
    };

    \addplot[
        color=lineran,
        mark=triangle*,
        mark size=1.75pt,
        line width=1.25pt,
        mark options={fill=lineran, draw=lineran}
    ] coordinates {
        (1,54.9)
        (2,57.7)
        (3,62.7)
        (4,66.8)
        (5,69.4)
    };

    \node[font=\footnotesize, anchor=east]
    at (axis description cs:0.02,-0.06) {Top-$k$};

    \end{axis}
\end{tikzpicture}
\end{tabular}%
}

\end{tabular}%
}
    \caption{
    Analysis of Memory Card Ordering and Selection.
    Fig.~\ref{fig:ordering_topk_a} shows that temporal ordering preserves the original event progression and provides a more coherent reasoning sequence after retrieval.
    Fig.~\ref{fig:ordering_topk_b} shows that retrieval-based selection is more robust across different Top-$k$ values than uniform or random selection.
    }
    \label{fig:ordering_topk}
    \vspace{-12pt}
\end{figure}

\textbf{Memory Card Ordering and Selection.}
Figure~\ref{fig:ordering_topk} examines two key decisions after Memory Card retrieval: how the selected cards should be ordered and how many cards should be selected for answer generation. 

Figure~\ref{fig:ordering_topk_a} shows that temporal ordering outperforms relevance-based and random ordering, indicating that chronological structure provides a more coherent reasoning sequence after relevant cards are selected. Figure~\ref{fig:ordering_topk_b} compares retrieval-based, uniform, and random selection under different Top-$k$ values. 
Retrieval is most effective under small Top-$k$, where the selected cards must be highly answer-relevant. 
As Top-$k$ increases, larger event coverage improves performance, but retrieval-based selection remains more stable than uniform or random selection. 
Full results are provided in Appendix~\ref{app:temporal_ordering_analysis} and Appendix~\ref{app:topk_retrieval_analysis}.

\section{Conclusion}
\label{sec:conclusion}

This paper presents \method{}, a video-memory-based augmentation framework for long-video QA. By constructing coherent semantic units and rendering representative visual moments with event-level video gists into unified Memory Cards, \method{} organizes sparse video clues into high-density multimodal evidence for VLM-based answering. This design shifts long-video QA from frame-centric evidence selection toward event-level evidence construction under constrained visual budgets. Experimental results show that \method{} improves multiple VLM backbones, while ablations confirm the effectiveness of semantic unit construction, Memory Card rendering, and temporal organization.

\section*{Limitations}

Although \method{} improves long-video QA, constructing Memory Cards introduces extra preprocessing cost. \method{} relies on a self-read VLM to segment videos into semantic units, generate event-level video gists, select representative visual moments, align speech transcripts, and render them into unified cards. These cards can be reused across questions from the same video, but the construction process still adds overhead before answering. Moreover, because each card uses a unit-level gist and representative visual moments, \method{} may be less effective for questions requiring fine-grained motion dynamics or continuous action understanding. Improving memory-construction efficiency while preserving dense event-level evidence remains an important direction for future work.

\section*{Acknowledgments}
This work is supported by the National Natural Science Foundation of China (No. 62576082 and No. 62461146205) and the Institute Guo Qiang at Tsinghua University.


\bibliography{custom}

\clearpage
\appendix

\section{Appendix}
\label{app:additional_results}

This appendix provides supplementary details and complete analysis results that complement the main paper.
We first describe the dataset licenses, dataset protocol, and controlled evaluation settings.
We then provide implementation details of Memory Card construction, retrieval, and answering.
Finally, we report additional analyses on component ablations, memory-session construction, fine-grained subtasks, resolution allocation, temporal ordering, and Top-$k$ sensitivity.

\subsection{License}
\label{app:license}

We use three publicly released long-video question answering benchmarks in this work: Video-MME~\cite{DBLP:conf/cvpr/FuDLLRZWZSZCLLZ25}, MLVU~\cite{zhou2024mlvu}, and LongVideoBench~\cite{DBLP:conf/nips/WuLCL24}. 
All datasets are used only for academic research and evaluation purposes, following their respective licenses and usage agreements. 
Video-MME is released for academic research use and prohibits commercial use. 
MLVU and LongVideoBench are released under the CC-BY-NC-SA-4.0 license. 
We do not redistribute the original videos or annotations, and all experiments are conducted on the officially released benchmark splits.

\subsection{Datasets and Evaluation Protocol}
\label{app:dataset_protocol}

We evaluate \method{} on three long-video question answering benchmarks: Video-MME, MLVU, and LongVideoBench. 
All benchmarks follow a multiple-choice evaluation protocol, and we report accuracy as the main metric.

\textbf{Video-MME.}
Video-MME~\cite{DBLP:conf/cvpr/FuDLLRZWZSZCLLZ25} evaluates comprehensive video understanding across diverse video durations and question categories. 
Following the setting in the main paper, we report overall accuracy and duration-wise accuracy on short, medium, and long videos.

\textbf{MLVU.}
MLVU~\cite{zhou2024mlvu} evaluates multi-task long-video understanding across diverse video genres and task types. 
We report the overall accuracy on the evaluated split.

\textbf{LongVideoBench.}
LongVideoBench~\cite{DBLP:conf/nips/WuLCL24} focuses on long-context video-language understanding. 
Following prior work, we evaluate on the validation split without interleaved subtitles and report overall accuracy.

\textbf{Controlled Evaluation.}
For controlled comparisons, we use Qwen2-VL, Qwen3-VL, and MiniCPM-V-4.5 as answering backbones. 
Unless otherwise specified, all variants use the same answering backbone, prompt format, decoding configuration, output protocol, answer extraction rule, and evaluation script. 
This protocol isolates the effect of the proposed Memory Card representation, retrieval strategy, ordering strategy, and resolution allocation without changing the answering VLM.

\subsection{Compared Methods}
\label{app:compared_methods}

Table~\ref{tab:main_results} compares \method{} with representative video VLMs and recent efficient long-video understanding methods. 
We briefly summarize them below.

\textbf{Video-LLaVA} aligns video representations with language models for video instruction following.
\textbf{Qwen-VL} is a general vision-language model used as an early multimodal baseline.
\textbf{VideoChat2} is a video instruction-tuned model for video dialogue and understanding.
\textbf{ShareGPT4Video} improves video understanding through high-quality video captions and instruction data.
\textbf{MovieChat} compresses dense frame tokens into sparse short-term and long-term memories for long-video understanding.
\textbf{MovieChat+} introduces question-aware memory consolidation to retain more information from question-relevant video segments.
\textbf{Video-XL} enables efficient long-video processing with substantially more input frames.
\textbf{LLaVA-OneVision} is a strong general-purpose VLM that transfers visual understanding across image and video tasks.
\textbf{Video-CCAM} enhances video-language understanding with causal cross-attention modeling.
\textbf{Frame-Voyager} learns to select informative frames for compact video understanding.
\textbf{LongVU} improves long-video understanding through visual-token compression and long-context processing.
For controlled comparisons, we evaluate Qwen2-VL, Qwen3-VL, and MiniCPM-V-4.5 together with their video-input variants, allowing us to isolate the effect of Memory Cards under the same answering backbones.

\subsection{Implementation Details}
\label{app:implementation_details}
\textbf{Memory Card Construction.}
For each video, \method{} first extracts multimodal clues for Memory Card construction.
We extract the audio track and use Qwen3-ASR to obtain timestamped speech transcripts.
Qwen3-VL-8B is used as the self-read VLM to segment each video into semantic sessions according to visual content and temporal event structure.
The speech transcripts are aligned to the resulting semantic sessions by timestamps and used as the spoken content of each session.
For videos without valid speech content, \method{} constructs Memory Cards using visual information only.

For each semantic session, the VLM generates an event-level video gist, where the topic serves as a compact session summary and the aligned speech transcript preserves spoken content when available.
The temporal span is retained as session metadata.
Representative visual moments are selected from each semantic session and rendered together with the corresponding gist and temporal metadata into image-based Memory Cards.

\textbf{Retrieval and Answering.}
For question answering, we use LongCLIP as the CLIP-style retriever to select question-relevant cards from the constructed Memory Card bank.
Memory Card construction is question-agnostic, while retrieval is question-aware.
The retrieved cards are assigned input resolutions according to retrieval relevance, reordered by their original temporal positions, and then fed into the answering VLM.
Unless otherwise specified, we use $k_{\mathrm{ret}}=44$, consisting of $4$ high-resolution, $8$ medium-resolution, and $32$ low-resolution Memory Cards.
Controlled comparisons use the same prompt format, decoding configuration, answer extraction rule, evaluation framework, and comparable visual input budget.

\begin{table}[t]
\centering
\small
\setlength{\tabcolsep}{6pt}
\renewcommand{\arraystretch}{1.2}
\begin{tabularx}{0.95\linewidth}{|>{\raggedright\arraybackslash}X|}
\hline
\rowcolor{promptblue}
\centering\textbf{Self-Read Prompt} \tabularnewline
\hline
\textbf{Prompt for semantic session Construction:} \\
{\ttfamily
You are a careful video analyst. You will read the entire video and divide it into semantically coherent sessions. Each session should correspond to a continuous event or topic in the video. Focus on the visual content and temporal event structure. For each session, provide its temporal span and concise topic summary. Do not use any downstream question, answer option, or ground-truth answer.
}
\\
\hline
\end{tabularx}
\caption{Prompt template used for self-read semantic session construction in \method{}. In implementation, the generated semantic sessions are associated with event-level video gists and temporal metadata for rendering Memory Cards.}
\vspace{-12pt}
\label{tab:selfread_prompt}
\end{table}

\textbf{Self-Read Prompt.}
To construct semantic sessions in a question-agnostic manner, we prompt the self-read VLM to analyze the entire video and divide it into temporally continuous events or topics. 
The prompt asks the VLM to focus on the visual content and temporal event structure, and to generate the temporal span and topic summary for each session. 
No downstream question, answer option, or ground-truth answer is used during this process. 
The resulting semantic sessions are associated with event-level video gists and temporal metadata for Memory Card rendering. 
The complete prompt template is shown in Table~\ref{tab:selfread_prompt}.

\subsection{Runtime Analysis}
\label{sec:runtime_analysis}

We analyze the computational cost of \method{} from two stages: one-time Memory Card construction and per-question inference. Table~\ref{tab:offline_runtime} reports the construction cost, while Table~\ref{tab:online_runtime} compares the online inference latency with direct-video inference.

\begin{table}[t]
\small
\centering
\setlength{\tabcolsep}{9pt}
\renewcommand{\arraystretch}{1.08}
\begin{tabular}{@{}lc@{}}
\toprule
\textbf{Construction Stage} & \textbf{Time / Video} \\
\midrule
ASR & 32.9 s \\
Semantic segmentation & 80.3 s \\
Gist + frame selection & 145.2 s \\
Card rendering & 4.7 s \\
\midrule
\textbf{Total} & \textbf{263.1 s} \\
\bottomrule
\end{tabular}
\caption{Average offline construction time per video on Video-MME.}
\label{tab:offline_runtime}
\end{table}

As shown in Table~\ref{tab:offline_runtime}, constructing the Memory Card bank takes 263.1 seconds per video on Video-MME. Most of the construction cost comes from semantic segmentation and event-level gist generation with representative-frame selection, while card rendering introduces relatively little overhead. This construction is performed once per video, after which the resulting Memory Card bank can be reused across downstream questions.

\begin{table}[t]
\small
\centering
\setlength{\tabcolsep}{7pt}
\renewcommand{\arraystretch}{1.08}
\begin{tabular}{@{}lccc@{}}
\toprule
\textbf{Method} & \textbf{Retrieval} & \textbf{Answering} & \textbf{Total} \\
\midrule
Direct Video & -- & 39.7 s & 39.7 s \\
\textbf{MemoryCard} & 47.2 ms & 8.7 s & 8.75 s \\
\bottomrule
\end{tabular}
\caption{Online inference time on Video-MME with Qwen3-VL-8B as the answering backbone.}
\label{tab:online_runtime}
\vspace{-4pt}

\end{table}
\begin{table*}[t]
\small
\centering
\setlength{\tabcolsep}{6pt}
\renewcommand{\arraystretch}{1.05}
\begin{tabular}{@{}lccc@{}}
\toprule
\textbf{Method} & \textbf{Video-MME} & \textbf{MLVU} & \textbf{LongVideoBench} \\
\midrule
Raw-frame Retrieval & 60.8 & 64.9 & 57.2 \\
\quad + Raw ASR & 62.2 & 65.3 & 58.2 \\
Unrendered MemoryCard & 62.6 & 65.5 & 58.6 \\
\textbf{MemoryCard} & \textbf{64.7} & \textbf{66.5} & \textbf{60.1} \\
\bottomrule
\end{tabular}
\caption{Comparison of transcript access, event-level evidence organization, and unified Memory Card rendering with Qwen3-VL.}
\label{tab:asr_rendering_ablation}
\end{table*}

\begin{table*}[t]
\centering
\small
\setlength{\tabcolsep}{5pt}
\renewcommand{\arraystretch}{1.15}
\begin{tabular}{lcccccc}
\toprule
\multirow{2}{*}{\textbf{Method}} 
& \multicolumn{4}{c}{\textbf{Video-MME}} 
& \multirow{2}{*}{\textbf{MLVU}} 
& \multirow{2}{*}{\textbf{LongVideoBench}} \\
\cmidrule(lr){2-5}
& \textbf{Overall} 
& \textbf{Short} 
& \textbf{Medium} 
& \textbf{Long} 
& 
& \\
\midrule

Qwen2-VL & 53.7 & 65.0 & 50.7 & 45.3 & 56.9 & 53.5 \\
\rowcolor{memoryblue}\textbf{MemoryCard} 
& \textbf{60.5} & \textbf{69.7} & \textbf{61.0} & 50.9 & 65.7 & \textbf{58.4} \\
\quad w/ Uniform-frame Units 
& 59.1 & 68.1 & 57.3 & \textbf{51.8} & 64.6 & 55.9 \\
\quad w/ Fixed-length Units 
& 59.0 & 68.0 & 58.8 & 50.1 & \textbf{65.9} & 56.1 \\
\quad w/ Shot-based Units 
& 58.7 & 67.9 & 58.1 & 50.1 & 64.9 & 56.6 \\

\midrule

Qwen3-VL & 57.4 & 68.8 & 53.1 & 50.2 & 57.2 & 56.3 \\
\rowcolor{memoryblue}\textbf{MemoryCard} 
& \textbf{64.7} & 72.2 & \textbf{64.8} & 54.7 & \textbf{66.5} & 60.1 \\
\quad w/ Uniform-frame Units 
& 62.5 & 71.6 & 62.3 & 53.6 & 65.3 & 59.1 \\
\quad w/ Fixed-length Units 
& 63.1 & 71.9 & 63.3 & 54.0 & 66.2 & 58.8 \\
\quad w/ Shot-based Units 
& 63.4 & \textbf{72.4} & 62.9 & \textbf{54.9} & 65.9 & \textbf{61.1} \\

\midrule

MiniCPM-V-4.5 & 59.9 & 69.6 & 58.2 & 52.0 & 57.0 & 55.6 \\
\rowcolor{memoryblue}\textbf{MemoryCard} 
& \textbf{67.2} & \textbf{76.0} & \textbf{67.6} & 58.0 & 69.4 & \textbf{62.0} \\
\quad w/ Uniform-frame Units 
& 66.5 & 74.6 & 66.0 & 58.9 & 68.3 & 61.4 \\
\quad w/ Fixed-length Units 
& 66.4 & 74.2 & 64.0 & \textbf{61.1} & 69.0 & 61.7 \\
\quad w/ Shot-based Units 
& 66.7 & 75.7 & 66.7 & 57.8 & \textbf{70.4} & 61.6 \\

\bottomrule
\end{tabular}
\caption{Effect of different memory unit construction strategies on three video question answering benchmarks with different answering backbones. The best results within each answering backbone are highlighted in \textbf{bold}.}
\label{tab:ablation_units}

\end{table*}

Table~\ref{tab:online_runtime} further reports the per-question inference cost after the Memory Card bank has been constructed. Instead of repeatedly processing the full video, \method{} answers each question using a fixed set of retrieved Memory Cards, substantially reducing online inference latency, with retrieval itself introducing little additional overhead. However, this benefit depends on amortizing the one-time construction cost across multiple questions. Under the three-question-per-video Video-MME setting, direct-video inference remains cheaper in total latency, and the approximate break-even point is nine questions per video. Therefore, the current implementation is better suited to reusable long-video understanding than strict single-query real-time inference.

\subsection{Component Ablations and Controlled Comparisons}
\label{app:component_ablation_settings}

This section provides detailed settings for the component ablations in the main paper and additional controlled comparisons that further separate the effects of transcript access, event-level evidence organization, and unified card rendering.

\subsubsection{Component Ablation Settings}
\begin{table*}[t]
\centering
\small
\setlength{\tabcolsep}{7pt}
\renewcommand{\arraystretch}{1.15}

\definecolor{posred}{RGB}{255,0,0}
\definecolor{negblue}{RGB}{0,0,255}

\begin{tabular}{lcccccc}
\toprule
\multirow{2}{*}{\textbf{Model}} 
& \multicolumn{6}{c}{\textbf{Video-MME}} \\
\cmidrule(lr){2-7}
& \textbf{Perception}
& \textbf{Recognition}
& \textbf{OCR}
& \textbf{Counting}
& \textbf{Reasoning}
& \textbf{Information Synopsis} \\
\midrule

Qwen2-VL-Video & 62.8 & 53.0 & 56.8 & 30.2 & 51.0 & 65.6 \\

Qwen2-VL & 65.9 & 54.9 & 56.8 & 30.6 & 50.5 & 65.9 \\

\quad \textbf{+ MemoryCard} & \textcolor{posred}{66.6} & \textcolor{posred}{62.9} & \textcolor{posred}{64.0} & \textcolor{posred}{39.9} & \textcolor{posred}{56.7} & \textcolor{posred}{76.5} \\

\midrule

Qwen3-VL-Video & 64.3 & 52.4 & 54.7 & 29.5 & 52.8 & 68.1 \\

Qwen3-VL & 63.9 & 58.1 & 58.3 & 35.4 & 57.9 & 71.2 \\

\quad \textbf{+ MemoryCard} & \textcolor{posred}{77.8} & \textcolor{posred}{64.1} & \textcolor{posred}{64.0} & \textcolor{posred}{38.8} & \textcolor{posred}{61.3} & \textcolor{posred}{78.0} \\

\midrule

MiniCPM-V-4.5-Video & 67.2 & 59.1 & 59.7 & 40.3 & 58.1 & 73.4 \\

MiniCPM-V-4.5 & 65.9 & 60.7 & 59.0 & 40.3 & 58.8 & 75.2 \\

\quad \textbf{+ MemoryCard} & \textcolor{posred}{73.1} & \textcolor{posred}{68.8} & \textcolor{posred}{73.4} & \textcolor{posred}{42.5} & \textcolor{posred}{66.7} & \textcolor{posred}{82.0} \\

\bottomrule
\end{tabular}
\caption{Performance of different models on Video-MME subtasks. \textcolor{posred}{Red} fonts represent positive results compared to the corresponding baseline, and 
\textcolor{negblue}{blue} fonts represent negative results.}

\vspace{-4pt}
\label{tab:videomme_subtasks}
\end{table*}

The component ablation in the main paper examines whether the gain of \method{} comes from question-aware retrieval alone or from the proposed rendered Memory Card representation. We provide the detailed settings below.

\textbf{Raw-frame Retrieval.}
This variant retrieves raw frames from the original video using the same LongCLIP-based retriever and a comparable visual budget as \method{}. 
The retrieved frames are sorted by timestamp and passed to the answering VLM. 
This setting isolates the effect of question-aware retrieval without changing the evidence unit.

\textbf{\method{} w/o Speech Transcript.}
This variant removes the aligned speech transcript from each Memory Card. 
The representative visual moment, topic, and temporal span are kept unchanged. 
This setting tests the contribution of spoken information and verifies whether the multimodal gain comes partly from rendering speech-derived clues.

\textbf{\method{} w/o Topic.}
This variant removes the topic generated by the self-read VLM, while keeping the representative visual moment, temporal span, and aligned speech transcript unchanged. 
This setting tests whether the generated unit-level summary helps the answering VLM interpret the selected visual evidence.

\textbf{\method{} w/o Temporal Span.}
This variant removes the temporal span from each Memory Card, while keeping the representative visual moment, topic, and aligned speech transcript unchanged. 
This setting tests whether explicit temporal grounding helps organize the selected evidence.

\textbf{\method{}.}
This is the full method. 
Each Memory Card contains a representative visual moment, an event-level video gist, and temporal metadata when available. 
The event-level video gist consists of the VLM-generated topic and the aligned speech transcript. 
The cards are retrieved according to the question, assigned resolutions according to relevance, sorted by temporal order, and passed to the answering VLM.

\subsubsection{Additional Controlled Comparisons}

To further distinguish the contribution of transcript access from event-level evidence organization and unified card rendering, we introduce two additional controlled variants.

\textbf{Raw-frame Retrieval + Raw ASR.}
This variant uses the same retrieved raw frames as Raw-frame Retrieval and additionally provides the temporally aligned raw ASR transcripts as textual input, without topic generation, temporal metadata, or card rendering.

\textbf{Unrendered MemoryCard.}
This variant uses the same semantic sessions, representative frames, topic, aligned transcript, and temporal metadata as \method{}, but provides the visual and textual evidence separately rather than rendering them into a unified image.

\begin{table*}[t]
\centering
\small
\setlength{\tabcolsep}{7pt}
\renewcommand{\arraystretch}{1.15}

\definecolor{tablered}{RGB}{255,0,0}
\definecolor{tableblue}{RGB}{0,0,255}

\begin{tabular}{lccccccccc}
\toprule
\multirow{2}{*}{\textbf{Model}} 
& \multicolumn{9}{c}{\textbf{MLVU}} \\
\cmidrule(lr){2-10}
& \textbf{TR} 
& \textbf{AR} 
& \textbf{VS} 
& \textbf{NQA} 
& \textbf{ER} 
& \textbf{PQA} 
& \textbf{SSC} 
& \textbf{AO} 
& \textbf{AC} \\
\midrule

Qwen2-VL-Video & 78.4 & 64.0 & 0.0 & 64.5 & 50.6 & 55.1 & 0.0 & 44.8 & 26.2 \\
Qwen2-VL & 83.8 & 58.5 & 0.0 & 61.4 & 56.8 & 59.9 & 0.0 & 45.9 & 20.4 \\
\quad \textbf{+ MemoryCard} 
& \textcolor{tablered}{87.5} 
& \textcolor{tableblue}{54.5} 
& 0.0 
& \textcolor{tablered}{79.4} 
& \textcolor{tablered}{66.2} 
& \textcolor{tablered}{68.8} 
& 0.0 
& \textcolor{tablered}{50.2} 
& \textcolor{tablered}{38.3} \\

\midrule

Qwen3-VL-Video & 79.8 & 60.0 & 0.0 & 58.6 & 51.1 & 55.1 & 0.0 & 47.9 & 17.5 \\
Qwen3-VL & 86.3 & 69.5 & 0.0 & 42.3 & 52.3 & 61.4 & 0.0 & 36.7 & 15.5 \\
\quad \textbf{+ MemoryCard} 
& \textcolor{tableblue}{85.6} 
& \textcolor{tableblue}{63.5} 
& 0.0 
& \textcolor{tablered}{73.2} 
& \textcolor{tablered}{61.1} 
& \textcolor{tablered}{73.5} 
& 0.0 
& \textcolor{tablered}{55.6} 
& \textcolor{tablered}{41.7} \\

\midrule

MiniCPM-V-4.5-Video & 85.9 & 65.0 & 0.0 & 65.6 & 55.1 & 63.6 & 0.0 & 41.3 & 22.8 \\
MiniCPM-V-4.5 & 84.8 & 68.0 & 0.0 & 58.0 & 54.0 & 63.8 & 0.0 & 37.5 & 20.9 \\
\quad \textbf{+ MemoryCard} 
& \textcolor{tablered}{85.6} 
& \textcolor{tableblue}{61.0} 
& 0.0 
& \textcolor{tablered}{80.8} 
& \textcolor{tablered}{64.8} 
& \textcolor{tablered}{76.4} 
& 0.0 
& \textcolor{tablered}{51.4} 
& \textcolor{tablered}{49.5} \\

\bottomrule
\end{tabular}
\caption{Performance of different models on MLVU subtasks. \textcolor{tablered}{Red} fonts represent positive results compared to the baseline, and 
\textcolor{tableblue}{blue} fonts represent negative results.}
\vspace{-10pt}
\label{tab:mlvu_subtasks}
\end{table*}
\begin{table*}[t]
\centering
\small
\setlength{\tabcolsep}{4.2pt}
\renewcommand{\arraystretch}{1.15}

\resizebox{\textwidth}{!}{
\begin{tabular}{lccccccccccccccccc}
\toprule
\multirow{2}{*}{\textbf{Model}} 
& \multicolumn{17}{c}{\textbf{LongVideoBench}} \\
\cmidrule(lr){2-18}
& \textbf{TOS} 
& \textbf{T2A} 
& \textbf{T2O} 
& \textbf{O2E} 
& \textbf{S2E} 
& \textbf{T3E} 
& \textbf{TAA} 
& \textbf{E3E} 
& \textbf{T3O} 
& \textbf{SSS} 
& \textbf{SOS} 
& \textbf{E2O} 
& \textbf{S2A} 
& \textbf{SAA} 
& \textbf{O3O} 
& \textbf{S2O} 
& \textbf{T2E} \\
\midrule

Qwen2-VL-Video
& 39.7 & 49.4 & 48.7 & 56.3 & 66.7 & 39.7 & 47.6 & 59.6 & 45.9 & 35.1 & 59.3 & 63.1 & 56.8 & 51.4 & 40.9 & 51.4 & 52.3 \\

Qwen2-VL
& 31.5 & 51.9 & 60.5 & 59.8 & 61.3 & 41.1 & 47.6 & 60.6 & 48.6 & 36.1 & 60.5 & 78.5 & 53.4 & 52.8 & 54.5 & 55.6 & 58.5 \\

\quad \textbf{+ MemoryCard}
& \textcolor{red}{42.6} & \textcolor{red}{63.8} & \textcolor{red}{64.9} & \textcolor{red}{64.7} & \textcolor{red}{73.3} & \textcolor{red}{42.6} & \textcolor{red}{56.7} & \textcolor{red}{68.8} & \textcolor{red}{52.4} & \textcolor{blue}{34.0} & \textcolor{red}{66.7} & \textcolor{blue}{70.2} & \textcolor{red}{75.5} & \textcolor{blue}{52.5} & \textcolor{red}{54.5} & \textcolor{red}{65.9} & \textcolor{red}{65.9} \\

\midrule

Qwen3-VL-Video
& 35.6 & 64.7 & 69.4 & 59.5 & 69.7 & 41.1 & 50.0 & 58.5 & 46.0 & 37.1 & 61.6 & 60.9 & 63.6 & 52.4 & 52.6 & 54.2 & 53.9 \\

Qwen3-VL
& 35.6 & 65.8 & 64.5 & 63.2 & 69.9 & 38.4 & 51.2 & 64.9 & 48.6 & 39.1 & 60.5 & 61.5 & 67.0 & 59.7 & 57.6 & 51.4 & 53.8 \\

\quad \textbf{+ MemoryCard}
& \textcolor{blue}{34.3} & \textcolor{red}{69.6} & \textcolor{red}{73.7} & \textcolor{red}{67.8} & \textcolor{red}{71.0} & \textcolor{red}{42.5} & \textcolor{red}{52.4} & \textcolor{blue}{58.5} & \textcolor{blue}{40.5} & \textcolor{red}{39.2} & \textcolor{red}{63.0} & \textcolor{red}{66.2} & \textcolor{red}{76.1} & \textcolor{blue}{58.3} & \textcolor{blue}{56.1} & \textcolor{red}{66.7} & \textcolor{red}{67.7} \\

\midrule

MiniCPM-V-4.5-Video 
& 41.1 & 49.4 & 65.8 & 64.4 & 67.7 & 43.8 & 46.3 & 69.2 & 54.1 & 40.2 & 63.0 & 60.0 & 68.2 & 47.2 & 57.6 & 55.6 & 60.0 \\

MiniCPM-V-4.5 
& 37.0 & 55.7 & 64.5 & 62.1 & 64.5 & 50.7 & 42.7 & 64.9 & 46.0 & 35.1 & 66.7 & 64.6 & 67.1 & 52.8 & 56.1 & 52.8 & 61.5 \\

\quad \textbf{+ MemoryCard} 
& \textcolor{red}{42.5} & \textcolor{red}{68.4} & \textcolor{red}{68.4} & \textcolor{red}{66.7} & \textcolor{red}{74.2} & \textcolor{red}{53.4} & \textcolor{red}{51.2} & \textcolor{red}{69.2} & \textcolor{red}{54.1} & \textcolor{red}{39.2} & \textcolor{red}{72.8} & \textcolor{red}{76.9} & \textcolor{red}{77.3} & \textcolor{red}{56.9} & \textcolor{blue}{54.6} & \textcolor{red}{70.8} & \textcolor{blue}{55.4} \\

\bottomrule
\end{tabular}
}
\caption{Performance of different models on LongVideoBench subtasks. \textcolor{red}{Red} fonts represent positive results compared to the baseline, and 
\textcolor{blue}{blue} fonts represent negative results.}
\vspace{-10pt}
\label{tab:longvideobench_subtasks}
\end{table*}

As shown in Table~\ref{tab:asr_rendering_ablation}, adding raw ASR transcripts to raw-frame retrieval consistently improves performance, confirming that speech information contributes to the overall gain. Unrendered MemoryCard further improves over the frame-plus-transcript setting, showing that event-level evidence organization provides benefits beyond transcript access alone. Finally, \method{} consistently outperforms the unrendered variant while using the same underlying visual and textual evidence, supporting the additional benefit of unified Memory Card rendering. Together, these results show that the gain of \method{} arises from the complementary effects of transcript access, event-level evidence organization, and unified rendering.

\subsection{Memory Session Construction Analysis}
\label{app:unit_construction_analysis}

\textbf{Setting.}
This analysis studies whether the quality of the source session affects the final Memory Card representation. 
All variants use the same rendering template, retriever, answering backbone, prompt format, visual budget, and evaluation protocol. 
The only difference is how the source sessions are constructed before rendering. 
Table~\ref{tab:ablation_units} reports the complete results for different memory-session construction strategies. \textbf{Uniform-frame Units} uniformly sample frames from the whole video and organize them into cards according to temporal order. 
\textbf{Fixed-length Units} divide each video into fixed temporal windows. 
\textbf{Shot-based Units} construct units according to visual shot or scene boundaries. 
The full \method{} uses self-read semantic sessions, where a VLM organizes the video into temporally coherent local events.

\textbf{Analysis.}
Self-read semantic sessions achieve strong overall performance and provide a reliable default source-session construction across answering backbones. 
This indicates that the benefit of \method{} does not only come from rendering information into image-based cards; the source session being rendered also matters. 
Uniform-frame and fixed-length units provide temporal coverage, but they may split a coherent event or merge unrelated content. 
Shot-based units capture low-level visual transitions, but shot boundaries do not always correspond to semantic event boundaries. 
In contrast, self-read semantic sessions better align with event-level video structure, making each Memory Card more self-contained and more suitable for retrieval and answering. 
This supports the design choice that Memory Cards should be built from semantic sessions rather than mechanically segmented clips.

\begin{table*}[t]
\centering
\small
\setlength{\tabcolsep}{3.2pt}
\renewcommand{\arraystretch}{1.12}
\resizebox{\textwidth}{!}{
\begin{tabular}{llccccccccccc}
\toprule
\multirow{2}{*}{\textbf{Backbone}}
& \multirow{2}{*}{\textbf{Setting}}
& \multicolumn{3}{c}{\textbf{Resolution Allocation}}
& \multirow{2}{*}{\textbf{\#Cards}}
& \multirow{2}{*}{\makecell{\textbf{Visual Budget}\\\textbf{/ Video}}}
& \multicolumn{4}{c}{\textbf{Video-MME}}
& \multirow{2}{*}{\textbf{MLVU}}
& \multirow{2}{*}{\makecell{\textbf{LongVideo}\\\textbf{Bench}}} \\
\cmidrule(lr){3-5}
\cmidrule(lr){8-11}
& 
& \textbf{High} & \textbf{Mid} & \textbf{Low}
& &
& \textbf{Overall} & \textbf{Short} & \textbf{Medium} & \textbf{Long}
& & \\
\midrule

\multirow{6}{*}{Qwen2-VL}
& R1 & 8 & 0 & 0 
& 8 
& 2,760
& 58.7 & 67.7 & 58.2 & 50.1 & 65.5 & 55.1 \\

& R2 & 6 & 6 & 8 
& 20 
& $2,790^{+1.1\%}$
& 59.9 & 68.8 & 59.4 & 51.6 & \textbf{65.8} & 55.2 \\

& R3 & 6 & 4 & 16 
& 26 
& $2,806^{+1.7\%}$
& 59.1 & 68.6 & 57.4 & 51.4 & 65.5 & 55.7 \\

& \mcblue{\textbf{R4 / Ours}} 
& \mcblue{\textbf{4}} & \mcblue{\textbf{8}} & \mcblue{\textbf{32}} 
& \mcblue{\textbf{44}} 
& \mcblue{$\textbf{2,852}^{\textbf{+3.3\%}}$}
& \mcblue{\textbf{60.5}} & \mcblue{69.7} & \mcblue{\textbf{61.0}} & \mcblue{50.9} & \mcblue{65.7} & \mcblue{\textbf{58.4}} \\

& R5 & 4 & 6 & 40 
& 50 
& $2,868^{+3.9\%}$
& 60.4 & 69.7 & 59.3 & \textbf{52.2} & 65.4 & 56.6 \\

& R6 & 4 & 4 & 48 
& 56 
& $2,884^{+4.5\%}$
& 59.7 & \textbf{70.3} & 58.3 & 50.4 & 65.5 & 56.2 \\

\midrule

\multirow{6}{*}{Qwen3-VL}
& R1 & 8 & 0 & 0 
& 8 
& 2,760
& 60.4 & 69.9 & 60.7 & 50.7 & 63.5 & 56.6 \\

& R2 & 6 & 6 & 8 
& 20 
& $2,790^{+1.1\%}$
& 62.1 & 70.8 & 62.1 & 53.6 & 66.0 & 57.5 \\

& R3 & 6 & 4 & 16 
& 26 
& $2,806^{+1.7\%}$
& 62.6 & 71.2 & 61.9 & 54.6 & 66.1 & 58.5 \\

& \mcblue{\textbf{R4 / Ours}} 
& \mcblue{\textbf{4}} & \mcblue{\textbf{8}} & \mcblue{\textbf{32}} 
& \mcblue{\textbf{44}} 
& \mcblue{$\textbf{2,852}^{\textbf{+3.3\%}}$}
& \mcblue{\textbf{64.7}} & \mcblue{\textbf{72.2}} & \mcblue{\textbf{64.8}} & \mcblue{54.7} & \mcblue{66.5} & \mcblue{\textbf{60.1}} \\

& R5 & 4 & 6 & 40 
& 50 
& $2,868^{+3.9\%}$
& 63.4 & 71.2 & 63.9 & 55.1 & 66.5 & 58.2 \\

& R6 & 4 & 4 & 48 
& 56 
& $2,884^{+4.5\%}$
& 63.6 & 71.7 & 63.9 & \textbf{55.3} & \textbf{66.8} & 59.3 \\

\midrule

\multirow{6}{*}{MiniCPM-V-4.5}
& R1 & 8 & 0 & 0 
& 8 
& 2,760
& 63.3 & 72.1 & 62.7 & 55.1 & 67.6 & 60.8 \\

& R2 & 6 & 6 & 8 
& 20 
& $2,790^{+1.1\%}$
& 65.1 & 73.2 & 64.4 & 57.8 & 68.9 & 61.0 \\

& R3 & 6 & 4 & 16 
& 26 
& $2,806^{+1.7\%}$
& 65.9 & 74.3 & 66.4 & 56.9 & 69.3 & 61.7 \\

& \mcblue{\textbf{R4 / Ours}} 
& \mcblue{\textbf{4}} & \mcblue{\textbf{8}} & \mcblue{\textbf{32}} 
& \mcblue{\textbf{44}} 
& \mcblue{$\textbf{2,852}^{\textbf{+3.3\%}}$}
& \mcblue{67.2} & \mcblue{76.0} & \mcblue{67.6} & \mcblue{\textbf{58.0}} & \mcblue{69.4} & \mcblue{\textbf{62.0}} \\

& R5 & 4 & 6 & 40 
& 50 
& $2,868^{+3.9\%}$
& \textbf{67.3} & \textbf{76.3} & \textbf{68.2} & 57.2 & 69.4 & 61.2 \\

& R6 & 4 & 4 & 48 
& 56 
& $2,884^{+4.5\%}$
& 67.1 & 76.1 & 67.8 & 57.4 & \textbf{70.2} &  61.0 \\

\bottomrule
\end{tabular}
}
\caption{Full results of resolution-allocation analysis under comparable visual budgets across different answering backbones. High, Mid, and Low denote the numbers of Memory Cards assigned to high, medium, and low resolutions, respectively. Visual Budget / Video reports the estimated visual patch/token cost computed by Eq.~\ref{eq:visual_budget}, with superscripts indicating the relative increase over R1.}
\vspace{-10pt}
\label{tab:4_resolution_allocation_analysis}
\end{table*}

\subsection{Fine-Grained Subtask Analysis}
\label{app:subtask_analysis}

\textbf{Setting.}
We further report fine-grained task-wise results on Video-MME, MLVU, and LongVideoBench. 
These results complement the overall accuracy by showing which types of questions benefit from the Memory Card representation. 
For Video-MME, we report six categories: perception, recognition, OCR, counting, reasoning, and information synopsis. 
For MLVU and LongVideoBench, we follow the task definitions of the corresponding benchmarks. 
Tables~\ref{tab:videomme_subtasks}, \ref{tab:mlvu_subtasks}, and~\ref{tab:longvideobench_subtasks} report the complete subtask results on the three benchmarks.

\textbf{Video-MME Analysis.}
The Video-MME subtask results show that \method{} improves multiple question types rather than a single isolated capability. 
The gains on perception, recognition, and OCR indicate that Memory Cards preserve fine-grained visual details from representative visual moments. 
The gains on reasoning and information synopsis suggest that the rendered event-level video gist helps the model interpret visual clues within a broader temporal and semantic structure.

Counting remains relatively challenging. 
This is reasonable because counting often requires exhaustive coverage over long temporal ranges, while \method{} follows a retrieve-then-answer pipeline and depends on the selected card set. 
Thus, the subtask results show both the strength and limitation of the current design: Memory Cards improve the density and interpretability of retrieved clues, but complete temporal enumeration remains difficult.

\textbf{MLVU Analysis.}
The MLVU results show that \method{} is especially useful for tasks that require event-level understanding, narrative context, or association between visual moments and surrounding information. 
This matches the intended role of Memory Cards: each card provides a representative visual moment together with its event-level video gist and temporal grounding.

Some categories show smaller or mixed gains. 
This suggests that tasks requiring fine-grained motion modeling or highly specialized action discrimination may not be fully solved by static card representations alone. 
Nevertheless, the overall improvement across backbones indicates that the event-level Memory Card bank provides a more useful input representation than isolated frames for long-video QA.

\textbf{LongVideoBench Analysis.}
The subtask results show that \method{} improves many temporal, object-centric, and state-association categories. 
These tasks require the model to locate relevant clues in long videos and reason with sufficient context, matching the intended role of Memory Cards.

The gains are not uniform across every subtask and backbone. 
This reveals a natural limitation of the current retrieve-then-answer framework: if retrieval misses a necessary card, or if the answering VLM fails to compare multiple selected cards correctly, the final answer may still be incorrect. 
Even so, the broad improvements demonstrate that structured Memory Cards provide more effective clue units for long-video understanding than sparse raw frames.

\textbf{Overall Observation.}
Across the fine-grained analyses, \method{} improves tasks that require visual detail, speech-related information, temporal grounding, and event-level synthesis. 
This supports the main claim of the paper: replacing isolated frame-level inputs with high-density Memory Cards improves the quality of clues provided to the answering VLM, while preserving compatibility with standard retrieve-then-answer inference.

\begin{table*}[t]
\centering
\small
\setlength{\tabcolsep}{4.8pt}
\renewcommand{\arraystretch}{1.12}
\begin{tabular}{clcccccc}
\toprule
\multirow{2}{*}{\textbf{Backbone}} 
& \multirow{2}{*}{\textbf{Card Order}}
& \multicolumn{4}{c}{\textbf{Video-MME}}
& \multirow{2}{*}{\textbf{MLVU}}
& \multirow{2}{*}{\textbf{LongVideoBench}} \\
\cmidrule(lr){3-6}
& & \textbf{Overall} & \textbf{Short} & \textbf{Medium} & \textbf{Long} & & \\
\midrule

\rowcolor{memoryblue}
\cellcolor{white}\multirow{3}{*}{Qwen2-VL}
& Temporal (Ours)  & \textbf{60.5} & \textbf{69.7} & \textbf{61.0} & 50.9 & \textbf{65.7} & \textbf{58.4} \\
& Relevance & 58.4 & 68.2 & 56.6 & 50.6 & 63.4 & 55.6 \\
& Random    & 59.1 & 68.2 & 57.3 & \textbf{52.0} & 63.2 & 55.4  \\

\midrule

\rowcolor{memoryblue}
\cellcolor{white}\multirow{3}{*}{Qwen3-VL}
& Temporal (Ours)  & \textbf{64.7} & \textbf{72.2} & \textbf{64.8} & \textbf{54.7} & \textbf{66.5} & \textbf{60.1} \\
& Relevance & 61.9 & 71.9 & 61.1 & 52.9 & 65.7 & 57.1 \\
& Random    & 61.2 & 71.4 & 59.4 & 52.7 & 65.3 & 56.7 \\

\midrule

\rowcolor{memoryblue}
\cellcolor{white}\multirow{3}{*}{MiniCPM-V-4.5}
& Temporal (Ours)  & \textbf{67.2} & \textbf{76.0} & \textbf{67.6} & \textbf{58.0} & \textbf{69.4} & \textbf{62.0} \\
& Relevance & 64.6 &73.7 &	63.6 &	56.4 & 68.4 & 59.2\\
& Random & 64.7 & 75.2 & 63.4 &	55.6 & 67.9 & 59.0\\

\bottomrule
\end{tabular}
\caption{Temporal ordering ablation of \method{} on three video question answering benchmarks. Temporal denotes the default setting where retrieved Memory Cards are ordered by video time before answering. Relevance keeps the retrieval ranking, while Random shuffles the retrieved Memory Cards with a fixed seed.}
\vspace{-10pt}
\label{tab:ablation_card_order}
\end{table*}

\subsection{Resolution Allocation Analysis}
\label{app:resolution_allocation_analysis}

\textbf{Setting.}
This analysis studies how to allocate visual resolution after Memory Cards are retrieved. 
Table~\ref{tab:4_resolution_allocation_analysis} reports the complete resolution allocation results under comparable visual budgets.
All settings use comparable visual budgets but distribute the budget differently across high-, medium-, and low-resolution cards.

We estimate the visual budget as the total visual patch/token cost of the selected Memory Cards. 
Let $N_h$, $N_m$, and $N_l$ denote the numbers of Memory Cards assigned to high, medium, and low resolutions, respectively, and let $C_h$, $C_m$, and $C_l$ denote the corresponding per-card visual costs after image preprocessing. 
The visual budget per video is computed as:
\begin{equation}
    B = N_h C_h + N_m C_m + N_l C_l .
    \label{eq:visual_budget}
\end{equation}
The relative increase reported in Table~\ref{tab:4_resolution_allocation_analysis} is computed with respect to R1:
\begin{equation}
    \Delta B = \frac{B - B_{\mathrm{R1}}}{B_{\mathrm{R1}}} \times 100\%.
\end{equation}

High-resolution cards preserve fine-grained visual details, while low-resolution cards provide broader temporal coverage with lower visual cost.

R1 allocates the budget to a small number of high-resolution cards. 
R5 and R6 increase the number of low-resolution cards to improve coverage. 
R4 is the default setting of \method{}, which assigns high resolution to the most relevant cards and uses medium- and low-resolution cards to preserve additional event context.

\subsection{Temporal Ordering Analysis}
\label{app:temporal_ordering_analysis}

\textbf{Setting.}
This analysis studies how the order of retrieved Memory Cards affects the answering VLM. 
The retrieved card set is kept unchanged across different variants; only the input order before answering is modified. 
\textbf{Temporal} is the default setting of \method{}, where retrieved cards are sorted according to their original timestamps. 
\textbf{Relevance} keeps the retriever ranking as the input order. 
\textbf{Random} shuffles the retrieved cards with a fixed seed. 
Table~\ref{tab:ablation_card_order} reports the complete results for different Memory Card ordering strategies.

\textbf{Analysis.}
Temporal ordering generally performs better than relevance ordering and random ordering on overall metrics. 
This result shows that retrieval relevance and reasoning order should not be conflated. 
Retrieval scores are useful for selecting question-relevant Memory Cards, but the ranking produced by the retriever does not necessarily preserve the event progression of the original video.

By restoring temporal order, \method{} preserves the chronological structure needed for long-video reasoning. 
This is especially important when questions involve event progression, state changes, or relations between temporally separated clues. 
The ordering analysis therefore supports a key design of \method{}: use relevance for selection and resolution allocation, but use video time for the final clue sequence.

\subsection{Top-$k$ Sensitivity and Selection Robustness}
\label{app:topk_retrieval_analysis}

\textbf{Setting.}
This analysis studies how the number of selected Memory Cards and the selection strategy affect performance. 
Different from the default multi-resolution configuration, this experiment uses a fixed high-resolution setting to isolate the effect of card selection. 
We vary Top-$k$ and compare three selection strategies. 
\textbf{Random} selects $k$ Memory Cards from the Memory Card bank with a fixed seed. 
\textbf{Uniform} selects $k$ Memory Cards uniformly along the video timeline. 
\textbf{Retrieval} selects the top-$k$ Memory Cards according to question-card relevance. 
Table~\ref{tab:ablation_topk_retrieval} reports the complete results for Top-$k$ sensitivity and selection robustness.

\textbf{Analysis.}
This experiment separates two questions. 
First, Random and Uniform selection test whether the constructed Memory Card bank itself contains useful video-level clues. 
Second, Retrieval tests whether question-aware selection can further focus the answering VLM on the most useful event-level clues.

The results show that increasing Top-$k$ generally improves evidence coverage under the fixed high-resolution setting, while retrieval-based selection is especially beneficial when only a limited number of cards can be selected. 
Together, this analysis verifies that Memory Cards provide a useful clue bank, while retrieval further improves efficiency by selecting question-relevant cards.

\begin{table*}[t]
\centering
\footnotesize
\setlength{\heavyrulewidth}{0.08em}
\setlength{\lightrulewidth}{0.05em}
\setlength{\cmidrulewidth}{0.03em}
\setlength{\tabcolsep}{3.8pt}
\renewcommand{\arraystretch}{1.08}
\begin{tabular*}{\textwidth}{@{\extracolsep{\fill}}llcccccc@{}}
\toprule
\multirow{2}{*}{\textbf{Selection}}
& \multirow{2}{*}{\textbf{Top-$k$}}
& \multicolumn{4}{c}{\textbf{Video-MME}}
& \multirow{2}{*}{\textbf{MLVU}}
& \multirow{2}{*}{\textbf{LongVideoBench}} \\
\cmidrule(lr){3-6}
& &
\textbf{Overall}
& \textbf{Short}
& \textbf{Medium}
& \textbf{Long}
& & \\
\midrule

\rowcolor{black!5}
\multicolumn{8}{@{}l}{\textbf{Qwen2-VL}} \\
\multirow{5}{*}{Random}
& 4  & 54.3 & 61.6 & 52.2 & 49.2 & 55.2 & 51.2 \\
& 8  & 56.6 & 65.2 & 54.6 & 49.2 & 57.1 & 53.1 \\
& 16 & 59.3 & 68.9 & 58.1 & 51.0 & 59.7 & 53.7 \\
& 32 & 61.7 & 71.0 & 62.6 & 51.7 & 62.5 & 54.9 \\
& 64 & 63.2 & 72.0 & 64.1 & 53.6 & 64.4 & 55.3 \\
\cmidrule(lr){1-8}

\multirow{5}{*}{Uniform}
& 4  & 53.1 & 59.8 & 48.4 & 51.1 & 54.6 & 52.6 \\
& 8  & 58.1 & 67.0 & 55.1 & 52.1 & 57.5 & 53.8 \\
& 16 & 59.8 & 69.6 & 58.0 & 51.8 & 59.7 & 56.1 \\
& 32 & 62.3 & 71.9 & 62.2 & 52.9 & 62.8 & 56.2 \\
& 64 & 62.8 & 72.9 & 62.4 & 53.1 & 62.1 & 55.2 \\
\cmidrule(lr){1-8}

\multirow{5}{*}{Retrieval}
& 4  & 57.1 & 66.4 & 56.7 & 48.2 & 63.7 & 54.5 \\
& 8  & 58.7 & 67.7 & 58.2 & 50.1 & 65.5 & 55.1 \\
& 16 & 61.0 & 70.8 & 59.9 & 52.3 & 66.3 & 56.2 \\
& 32 & 62.5 & 72.8 & 61.2 & 53.4 & 67.2 & 57.4 \\
& 64 & 62.3 & 72.4 & 61.6 & 52.8 & 66.7 & 56.8 \\

\midrule

\rowcolor{black!5}
\multicolumn{8}{@{}l}{\textbf{Qwen3-VL}} \\
\multirow{5}{*}{Random}
& 4  & 54.9 & 62.2 & 51.8 & 50.6 & 47.9 & 51.0 \\
& 8  & 57.7 & 67.3 & 55.8 & 50.0 & 51.8 & 51.8 \\
& 16 & 62.7 & 73.7 & 61.2 & 53.3 & 56.7 & 55.9 \\
& 32 & 66.8 & 77.0 & 66.0 & 57.3 & 60.9 & 58.0 \\
& 64 & 69.4 & 79.6 & 70.4 & 58.1 & 67.1 & 61.7 \\
\cmidrule(lr){1-8}

\multirow{5}{*}{Uniform}
& 4  & 55.6 & 61.7 & 52.2 & 53.0 & 47.6 & 50.2 \\
& 8  & 60.2 & 70.9 & 57.1 & 52.6 & 52.4 & 54.4 \\
& 16 & 63.1 & 72.4 & 62.3 & 54.7 & 58.2 & 56.5 \\
& 32 & 66.7 & 75.6 & 66.8 & 57.8 & 63.8 & 59.5 \\
& 64 & 69.0 & 79.2 & 69.6 & 58.3 & 68.4 & 61.3 \\
\cmidrule(lr){1-8}

\multirow{5}{*}{Retrieval}
& 4  & 58.3 & 68.0 & 58.3 & 48.4 & 60.3 & 54.0 \\
& 8  & 60.4 & 69.9 & 60.7 & 50.7 & 63.5 & 56.6 \\
& 16 & 64.3 & 73.9 & 63.9 & 55.2 & 66.6 & 59.3 \\
& 32 & 68.1 & 78.0 & 68.4 & 57.9 & 67.3 & 60.4 \\
& 64 & 70.4 & 81.7 & 71.6 & 58.1 & 69.8 & 61.7 \\

\midrule

\rowcolor{black!5}
\multicolumn{8}{@{}l}{\textbf{MiniCPM-V-4.5}} \\
\multirow{5}{*}{Random}
& 4  & 59.0 & 67.0 & 56.4 & 53.7 & 54.0 & 54.1 \\
& 8  & 61.4 & 69.6 & 60.2 & 54.4 & 57.5 & 55.2 \\
& 16 & 65.3 & 75.8 & 64.3 & 55.8 & 61.9 & 57.1 \\
& 32 & 67.9 & 78.1 & 67.0 & 58.4 & 65.7 & 59.2 \\
& 64 & 68.7 & 80.1 & 68.8 & 57.2 & 69.9 & 59.6 \\
\cmidrule(lr){1-8}

\multirow{5}{*}{Uniform}
& 4  & 57.9 & 65.4 & 54.8 & 53.6 & 53.9 & 54.3 \\
& 8  & 62.8 & 73.6 & 59.4 & 55.3 & 58.7 & 56.1 \\
& 16 & 66.6 & 78.2 & 63.9 & 57.8 & 63.2 & 59.8 \\
& 32 & 69.1 & 78.8 & 69.1 & 59.4 & 67.2 & 60.5 \\
& 64 & 69.2 & 79.8 & 70.3 & 57.6 & 69.8 & 60.1 \\
\cmidrule(lr){1-8}

\multirow{5}{*}{Retrieval}
& 4  & 60.8 & 69.0 & 59.9 & 53.6 & 63.1 & 56.9 \\
& 8  & 63.3 & 72.1 & 62.7 & 55.1 & 66.2 & 59.7 \\
& 16 & 65.9 & 74.2 & 64.8 & 58.7 & 68.0 & 60.8 \\
& 32 & 68.1 & 77.6 & 68.1 & 58.8 & 70.6 & 61.9 \\
& 64 & 69.3 & 77.4 & 70.1 & 60.3 & 71.1 & 59.6 \\

\bottomrule
\end{tabular*}
\caption{Top-$k$ sensitivity and Memory Card selection robustness under a fixed high-resolution setting. For all variants, the selected $k$ Memory Cards are fed to the backbone at high resolution. Random, Uniform, and Retrieval denote random sampling, temporal uniform sampling, and question-card relevance retrieval, respectively.}
\vspace{-10pt}
\label{tab:ablation_topk_retrieval}
\end{table*}

\FloatBarrier

\subsection{Case Study}

\begin{figure*}[t]
\includegraphics[width=\textwidth]{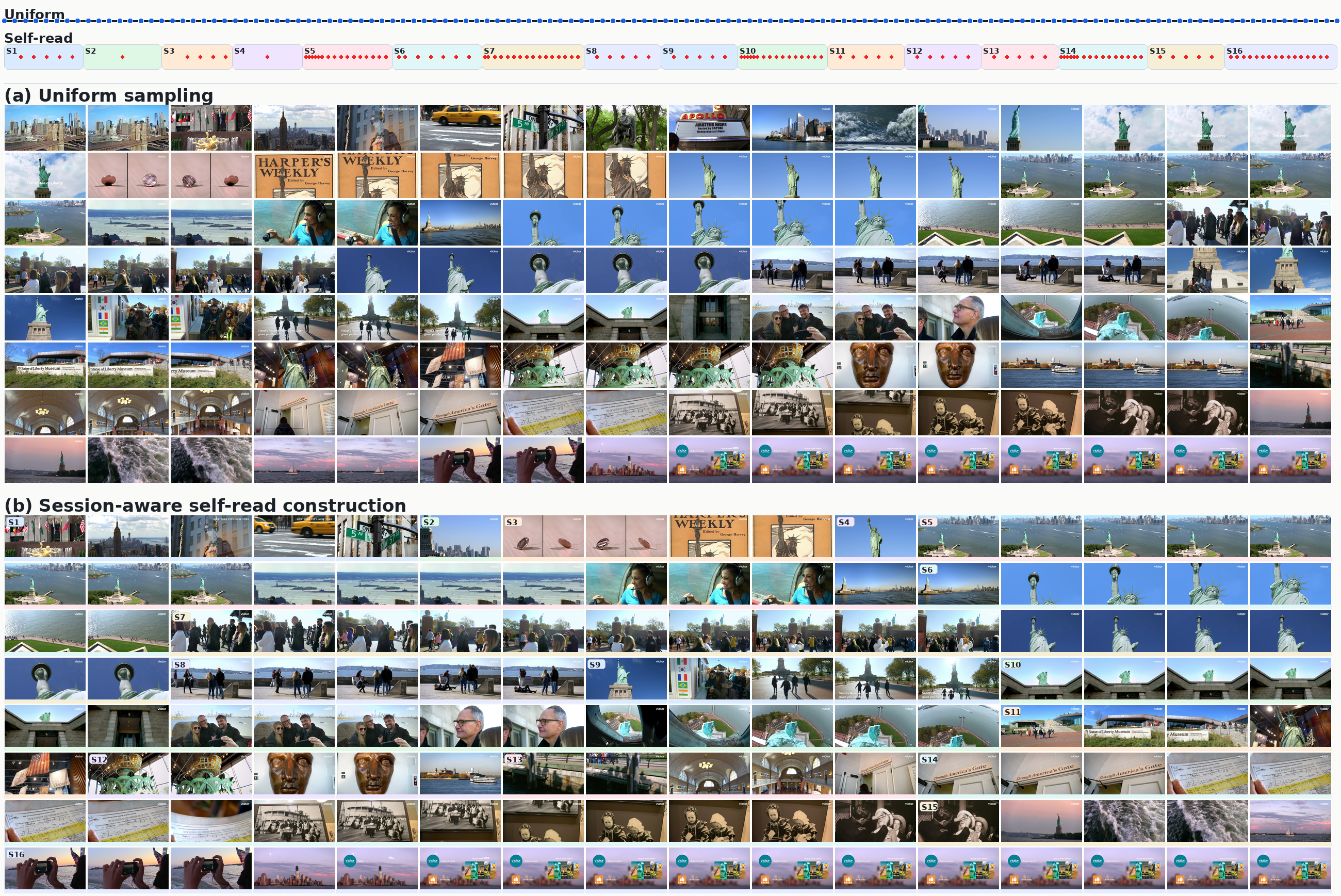}
\caption{
Uniform Sampling vs. Session-Aware Self-Read Construction under the same 128-frame visual budget.
Uniform sampling selects frames at fixed global intervals, while self-read first segments the video into semantic sessions and then selects keyframes within each session.
Colored blocks indicate session boundaries, highlighting that our visual evidence is organized by event structure rather than uniform temporal spacing.
}
\label{fig:uniform_vs_selfread_case}
\vspace{-16pt}
\end{figure*}

We provide qualitative examples to illustrate how \method{} constructs and uses Memory Cards for long-video question answering. 
Figure~\ref{fig:uniform_vs_selfread_case} compares uniform sampling with self-read semantic-unit construction under the same visual budget. 
Uniform sampling distributes frames at fixed temporal intervals, while self-read construction organizes the video according to event structure and selects representative visual moments within each semantic session. 
This produces evidence that is better aligned with meaningful video content rather than raw timestamps.

\begin{figure*}[t]
\includegraphics[width=\textwidth]{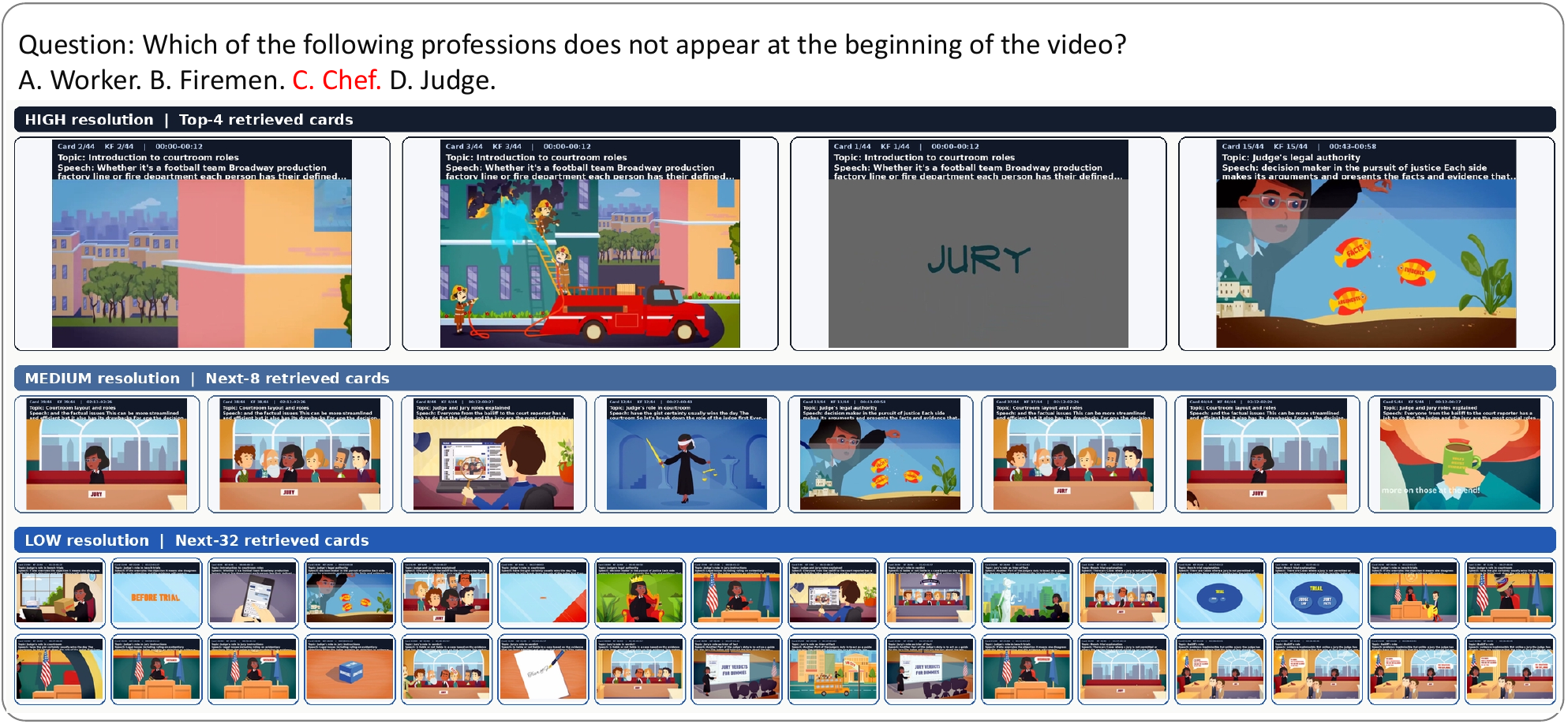}
\caption{
Question-Conditioned Retrieval Visualization.
For each multiple-choice question, \method{} retrieves Memory Cards from the self-read memory bank and organizes them into 4 high-resolution, 8 medium-resolution, and 32 low-resolution cards.
The ground-truth option is highlighted in red.
The retrieved cards serve as the visual evidence provided to the answering VLM.
}
\label{fig:case_question_retrieval_1}
\vspace{-16pt}
\end{figure*}

\begin{figure*}[t]
\includegraphics[width=\textwidth]{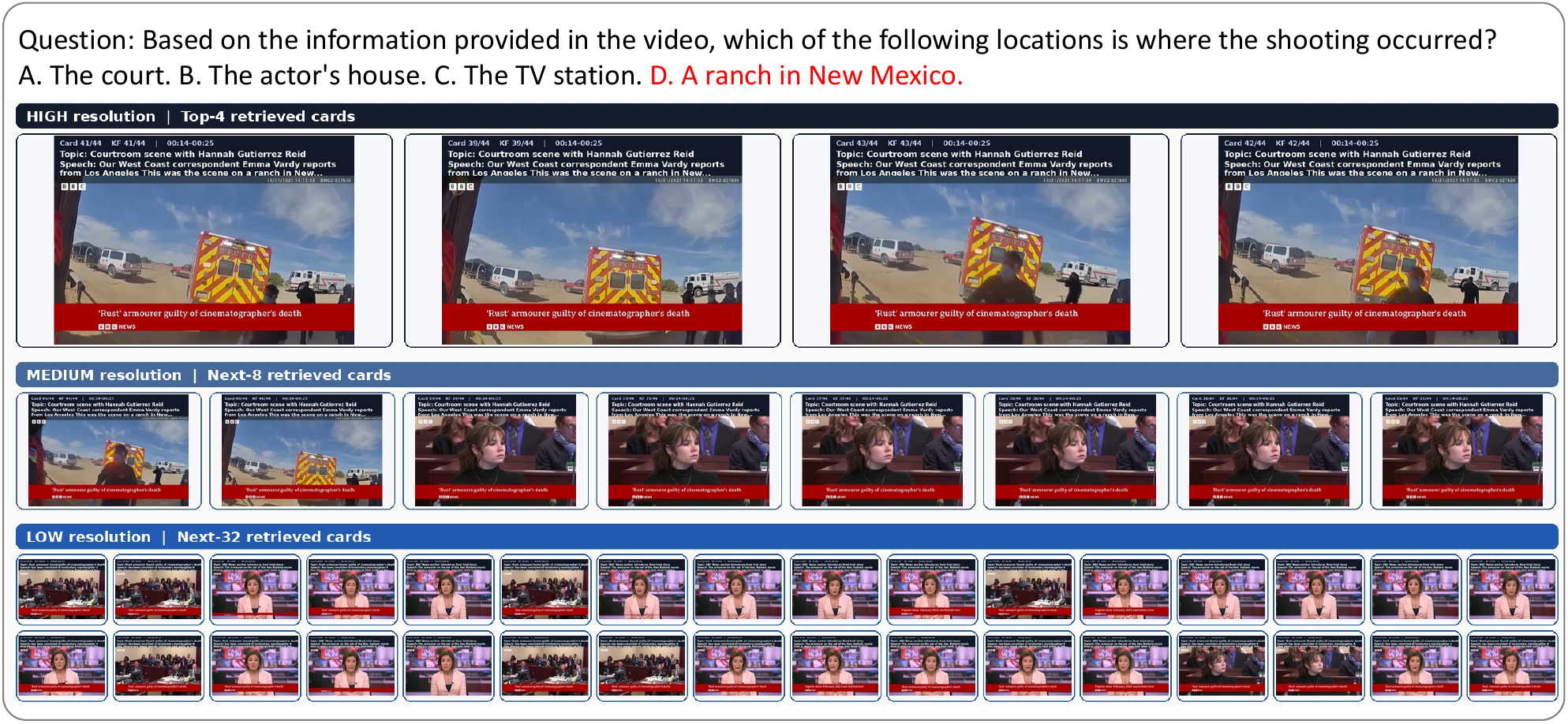}
\caption{
Question-Conditioned Retrieval Visualization.
For each multiple-choice question, \method{} retrieves Memory Cards from the self-read memory bank and organizes them into 4 high-resolution, 8 medium-resolution, and 32 low-resolution cards.
The ground-truth option is highlighted in red.
The retrieved cards serve as the visual evidence provided to the answering VLM.
}
\label{fig:case_question_retrieval_2}
\vspace{-16pt}
\end{figure*}

Figures~\ref{fig:case_question_retrieval_1} and~\ref{fig:case_question_retrieval_2} further illustrate how \method{} uses the constructed Memory Card bank for question-conditioned retrieval. 
Given a multiple-choice question, \method{} retrieves the most relevant Memory Cards from the question-independent memory bank, assigns higher resolutions to more relevant cards and lower resolutions to contextual cards, and then restores their original temporal order before answer generation. 
The retrieved cards therefore preserve fine-grained visual details for answer-critical events while retaining broader event-level semantic, speech, and temporal context across the video. 
These examples qualitatively show how \method{} converts sparse and temporally dispersed clues into a compact, coherent evidence set for the answering VLM, complementing the quantitative improvements observed in the main experiments.

\clearpage

\end{document}